\documentclass[journal]{IEEEtran}
\usepackage{cite}
\usepackage{amsmath,amssymb,amsfonts}
\usepackage{algorithmic}
\usepackage{graphicx}
\usepackage{textcomp}
\usepackage{tabularx,array}
\usepackage{url}
\usepackage{adjustbox}
\usepackage{bm}
\def\BibTeX{{\rm B\kern-.05em{\sc i\kern-.025em b}\kern-.08em
    T\kern-.1667em\lower.7ex\hbox{E}\kern-.125emX}}

\begin{document}

\title{A Unified Dynamics Framework for Reinforcement Learning and Classical Control of a Six-DOF Pipeline-Tracking ROV in NVIDIA Isaac Sim}

\author{Cheng~Siong~Chin, M.~Venkateshkumar, and Jianhua~Zhang
\thanks{C. S. Chin is with the Faculty of Science, Agriculture, and Engineering, Newcastle University Singapore, Singapore 828608, Singapore (e-mail: cheng.chin@newcastle.ac.uk).}
\thanks{M. Venkateshkumar is with the Amrita School of Engineering, Coimbatore, Amrita Vishwa Vidyapeetham, India (e-mail: m\_venkateshkumar@cb.amrita.edu).}
\thanks{J. Zhang is with the School of Information and Control Engineering, Qingdao University of Technology, Qingdao 266525, China (e-mail: jianhuazhang@qut.edu.cn).}}

\markboth{Chin \textit{et al.}: Unified Fossen-Dynamics Framework for RL and Classical Control of a Pipeline-Tracking ROV}
{Chin \textit{et al.}: Unified Fossen-Dynamics Framework for RL and Classical Control of a Pipeline-Tracking ROV}

\maketitle

\begin{abstract}
Reinforcement learning controllers for underwater vehicles are usually trained against one physics representation and, when a live simulator or hardware is available, deployed against another, so reported tracking performance does not always describe what the controller will do once it leaves the training environment. This paper reports a pipeline-tracking control architecture for a six-degree-of-freedom remotely operated vehicle (ROV) in which a single Universal Scene Description (USD) scene supplies the real BlueROV2-Heavy mass, added-mass, damping, buoyancy, and eight-thruster allocation parameters to both halves of the system: a vectorized NumPy implementation of Fossen's marine-craft equations, and an interactive NVIDIA Isaac Sim deployment in which the identical equations are applied as forces to a PhysX rigid body at every physics step. The Coriolis-centripetal coupling term C(\ensuremath{\nu})\ensuremath{\nu} is included as the primary dynamics model in both branches, and a controlled ablation restricted to proximal policy optimization (PPO) and trust region policy optimization (TRPO) shows that including the term does not destabilize either algorithm and modestly improves tracking accuracy, an approximately 19 percent reduction in standoff root-mean-square error for TRPO. Five reinforcement learning algorithms, PPO, soft actor-critic (SAC), twin-delayed deep deterministic policy gradient (TD3), deep deterministic policy gradient (DDPG), and TRPO, are trained against an identical environment, reward, and 30-episode randomized evaluation harness through a checkpoint-compatibility layer that lets any trained policy be scored by the same evaluation code. The same shared-dynamics pipeline is extended with six classical and model-based control baselines, proportional-integral-derivative (PID) control, sliding-mode control, fuzzy logic control, feedback linearization, model predictive control, and an adaptive neuro-fuzzy inference system (ANFIS), each driven by the identical guidance geometry and thruster-allocation map as the learned policies, so that any performance difference reflects the control law itself. Under Coriolis-enabled dynamics, PPO, TRPO, and feedback linearization achieve the strongest combination of a 100 percent success rate and competitive-to-best tracking accuracy; PID, fuzzy logic control, and the ANFIS surrogate also reach 100 percent success with somewhat looser tracking; DDPG and TD3 each show a specific, explainable failure mode rather than a general weakness of off-policy learning; and classical, model-based control remains a strong, competitive, low-cost baseline against the best learned policies on this task.
\end{abstract}


\section{Introduction}

Subsea pipelines carry oil, gas, and slurry across seabeds that divers and tethered inspection vehicles cannot reach cheaply or safely, and the standard inspection task for a remotely operated vehicle (ROV) is to fly a fixed standoff distance above the pipeline while avoiding rock outcrops and the seafloor itself \cite{b1}. The six coupled degrees of freedom of an underwater vehicle, surge, sway, heave, roll, pitch, and yaw, make the task difficult to hand-code, since damping, added mass, Coriolis-centripetal coupling, and buoyancy interact nonlinearly with vehicle attitude \cite{b2}. Reinforcement learning (RL) offers an alternative to hand-tuned controllers: a policy trained against a physics model learns the coupling directly from reward signal rather than from a linearized approximation of it, and recent work reports reinforcement-learning controllers  \cite{b3,b4,b5} that track paths, hold depth, and dock autonomous underwater vehicles under disturbances that defeat a fixed-gain proportional-integral-derivative loop.

Most of this work is validated inside the same simulator used for training, with a single algorithm, and rarely against a classical or model-based controller built from the same dynamics model. A policy is trained and evaluated against one instance of a dynamics model, sometimes with domain randomization applied to the model's own parameters, and the paper reports how that policy performed inside the training loop \cite{b6,b7}. Whether the same policy behaves identically once handed to a different physics engine, whether a reported result generalizes across the family of algorithms used for continuous control, and whether a learned policy actually outperforms a well-tuned classical controller given the same dynamics model and guidance geometry, are typically left open \cite{b8}. NVIDIA Isaac Sim narrows the first question for many robot classes by pairing a GPU-accelerated PhysX engine with a Universal Scene Description (USD) scene graph that serves as a single source of truth for a robot's geometry and physical parameters \cite{b9,b10}, and a recent underwater-specific extension, MarineGym, adapts Isaac Sim's physics backbone to marine hydrodynamics at high simulation throughput \cite{b11}. A worked architecture in which the same, Coriolis-coupled equations of motion, read from one parameter set, drive a vectorized trainer, an interactive deployment scene, and a shared bank of classical baselines alike, has not been reported for a pipeline-tracking task.

This paper builds a USD scene encoding a 6-DOF ROV's mass, added mass, linear and quadratic damping, buoyancy, and eight-thruster allocation matrix as scene attributes, parsed by two independent code paths: a NumPy-vectorized implementation of Fossen's marine-craft dynamics \cite{b2}, with the Coriolis-centripetal coupling term C(\ensuremath{\nu})\ensuremath{\nu} included as part of the default governing equation, integrated with a fixed-step fourth-order Runge-Kutta scheme across 64 parallel environments; and an interactive Isaac Sim deployment in which a controller reads the live PhysX pose and velocity of the rigid body at every physics step, computes the identical Coriolis, damping, thruster, and buoyancy terms, and applies them as a body-frame force and torque, letting PhysX integrate the vehicle in place of a second Runge-Kutta solver. A controlled ablation, restricted to two of the five compared reinforcement-learning algorithms, shows that including this term is safe and modestly beneficial, motivating its use as the primary model throughout rather than as a simplification to be engineered around. Against this Coriolis-inclusive architecture, five reinforcement learning algorithms are derived, trained, and evaluated: PPO \cite{b12} with generalized advantage estimation \cite{b13}, SAC \cite{b14}, TD3 \cite{b15}, DDPG \cite{b16}, and TRPO \cite{b17}, scored through a single checkpoint-compatibility layer built around PPO's own evaluation code; and six classical and model-based controllers, PID, sliding-mode control, fuzzy logic control, feedback linearization, model predictive control, and an adaptive neuro-fuzzy inference system, are added as baselines under the identical guidance geometry and thruster-allocation map, so that any performance difference reflects the control law rather than the task setup. Sections III through V present the shared architecture, dynamics, and RL environment; Sections VI and VII derive the five reinforcement-learning algorithms; Section VIII presents the six classical and model-based baselines; Section IX describes the Isaac Sim deployment; Sections X and XI give the evaluation protocol and the resulting comparison; Section XII discusses the comparison and its limitations.

\section{Related Work}

\subsection{Guidance and control for underwater vehicles}

Line-of-sight guidance coupled with proportional-integral-derivative or sliding-mode control remains the reference approach for autonomous underwater vehicle (AUV) path following; a recent survey documents three decades of refinements to the guidance law, from constant look-ahead distance to adaptive and time-varying variants that cope with current disturbances \cite{b1}. These methods are computationally cheap and well understood, but each new disturbance model or actuator constraint typically requires a new stability proof, and the resulting controllers do not adapt online to dynamics they were not designed against.

\subsection{Reinforcement learning for underwater tracking and following}

Deep deterministic policy gradient and its twin-delayed variant have been used for AUV path following and depth tracking under current disturbance, with reported reductions in cross-track error of up to two-thirds relative to a PID baseline once a recurrent state encoder and a learnable guidance parameter are added \cite{b4}. Proximal policy optimization has been used for three-degree-of-freedom motion planning without a prior disturbance model \cite{b8}, and soft actor-critic variants, sometimes combined with a long short-term memory encoder, produce adaptive trajectory-tracking controllers for underwater vehicles \cite{b18,b19}. Obstacle-aware extensions add a real-time avoidance term to the reward or the action space, closer to the pipeline-tracking task addressed here, since a policy that only tracks a centerline still routes around fixed and moving obstacles \cite{b20}. Across this literature, a learned policy generalizes across current strength and payload changes that would require re-derivation of a classical controller; the recurring gap is that a policy is validated inside the dynamics model it was trained against, with sim-to-real or sim-to-sim transfer left as future work \cite{b3,b5}, that a single algorithm is typically reported per task, and that a classical baseline built from the same dynamics model is rarely reported alongside the learned policy.

\subsection{Off-policy and trust-region continuous control}

DDPG combines a deterministic policy with an actor-critic architecture and a replay buffer, extending value-based deep reinforcement learning to continuous action spaces, and is sensitive to function-approximation error in its single critic and to exploration-noise scale \cite{b16}. TD3 addresses this with twin critics under clipped double-Q learning, delayed policy updates, and target-policy smoothing noise, reporting more stable learning than DDPG on standard continuous-control benchmarks \cite{b15}. SAC frames the objective as maximum-entropy reinforcement learning, adding an entropy bonus to the reward and automatically tuning the temperature coefficient against a target entropy \cite{b14}. TRPO constrains each policy update to a trust region bounded by the Kullback-Leibler divergence between successive policies, computed via a conjugate-gradient approximation to the natural gradient and enforced with a backtracking line search, and is the on-policy predecessor PPO's clipped surrogate objective was designed to approximate more cheaply \cite{b17}. Comparisons spanning all of these families against one environment, reward, and evaluation code are less common than single-algorithm reports, which motivates the comparison undertaken here.

\subsection{Simulation platforms and the sim-to-real gap}

Purpose-built underwater simulators such as HoloOcean and UNav-Sim provide photorealistic rendering and sensor models for AUV research but were not designed for the throughput population-based reinforcement learning needs \cite{b21,b22}. General-purpose GPU-accelerated simulators built for terrestrial and aerial robots, most visibly Isaac Gym and its successor Isaac Sim, solved the throughput problem by vectorizing physics and observation computation on the GPU \cite{b9,b23}. Isaac Sim's Universal Scene Description graph lets a single scene file describe a robot's geometry, sensors, and physical parameters once, consumed by rendering, physics, and any downstream training or deployment script that opens the same file \cite{b10}. MarineGym extends Isaac Sim's PhysX backbone with a hydrodynamic plugin for underwater vehicles and a domain-randomization toolkit aimed at narrowing the sim-to-real gap \cite{b11}. Domain randomization itself, varying physical parameters during training so a policy does not overfit to one simulator instance, was established for manipulation and has since been applied to marine vehicles directly \cite{b24,b25}.

\subsection{The Omniverse, Isaac Sim, and Isaac Lab simulation stack}

Three related but distinct layers of NVIDIA's robotics-simulation stack (see Figure 1) are referred to throughout this paper and are worth separating explicitly. Omniverse is the underlying platform and engine: it provides the core 3D graphics, physics, and simulation infrastructure, together with the USD-based scene representation of Section III-A, on which higher-level tools are built. Isaac Sim is the robotics simulation environment built on Omniverse, supplying the virtual world, robots, sensors, physics, and robotics-specific tooling, including a ROS 2 bridge, needed to construct and test a vehicle such as the ROV modelled here. Isaac Lab is a further, reinforcement-learning-oriented framework built around Isaac Sim, providing a vectorized training environment and tooling for running many parallel simulation instances against a single policy-optimization loop \cite{b26}. An engine, in this sense, is the software infrastructure that performs the underlying computation, rendering, collision and force calculation, and standardized scene representation, so that applications built on top of it do not have to develop these capabilities from scratch, in the same sense that a car's engine supplies motive power a driver does not directly interact with.

A generic Isaac-centred workflow for this class of problem proceeds in four stages: building and validating a vehicle model and its hydrodynamic behaviour inside Isaac Sim's GUI and visual debugging tools; defining a reinforcement-learning task in Isaac Lab by specifying observations (vehicle pose, velocity, and target waypoint), actions (thruster commands), and a reward (reaching the target while minimizing energy consumption); training a control policy with many parallel environment instances running under Isaac Lab; and deploying the trained policy back into Isaac Sim for visual evaluation, optionally bridged to ROS 2 for robotics-system integration. Figure 2 illustrates this pipeline in the general case, including an optional connection to a physical vehicle.

\begin{figure}[!t]
\renewcommand{\thefigure}{2}
\centering
\includegraphics[width=\columnwidth,height=0.8\textheight,keepaspectratio]{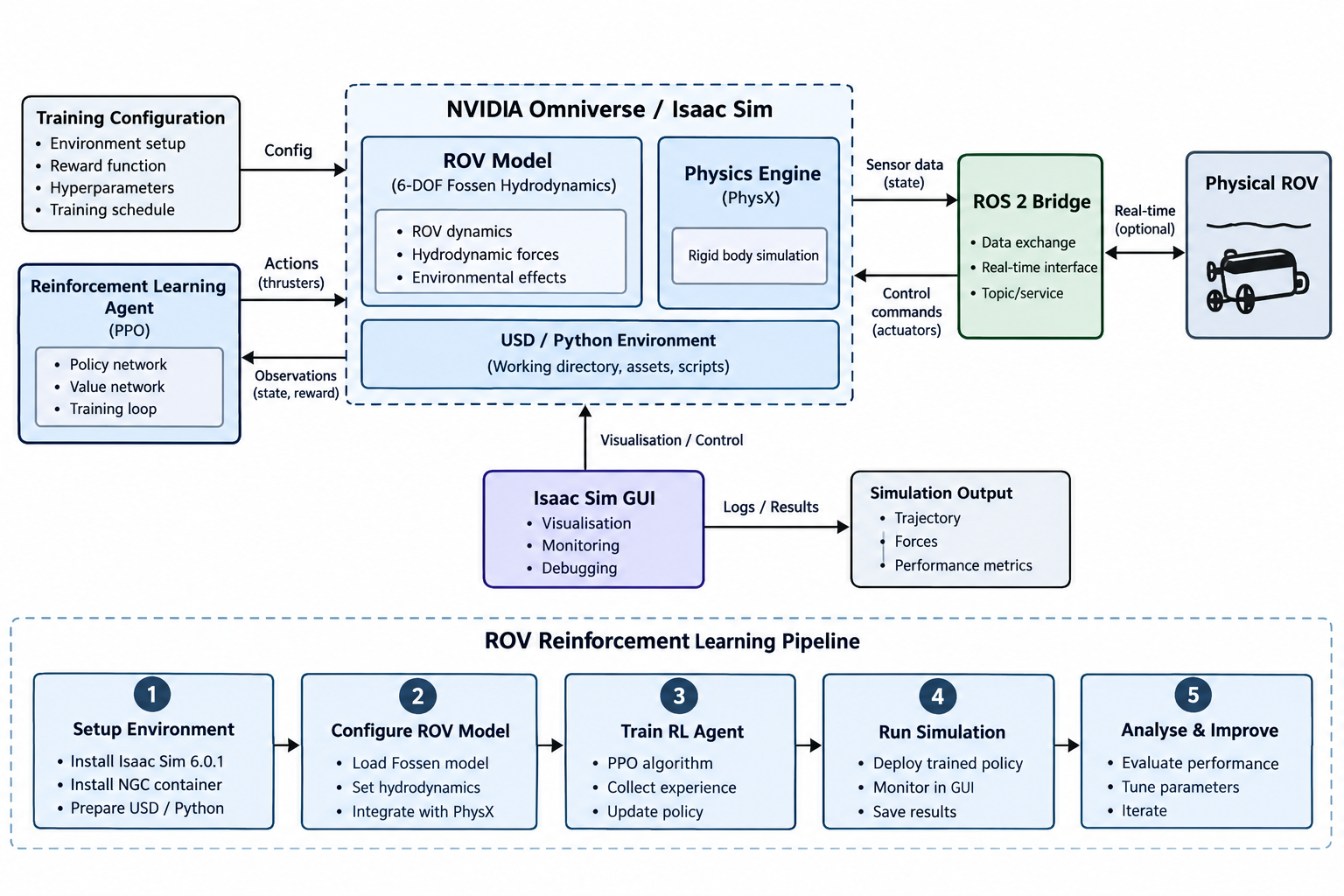}
\caption{Reference reinforcement-learning pipeline for a vehicle simulated in NVIDIA Omniverse / Isaac Sim, with an optional Isaac Lab training loop and an optional ROS 2 bridge to a physical vehicle (reference diagram).}
\label{fig:2}
\end{figure}

The architecture reported in this paper (Section III onward) follows the Isaac Sim half of this stack directly: the ROV model and its hydrodynamics are built and validated as a USD scene, and the trained policy is deployed and evaluated visually inside Isaac Sim, with an equivalent ROS 2 bridge available but not exercised. Training itself, however, is carried out by a purpose-written, NumPy-vectorized implementation of the same Fossen dynamics described in Section IV, so that the same five algorithms of Sections VI and VII, and the six classical baselines of Section VIII, can be trained, implemented, and compared at a wall-clock cost of minutes rather than hours on the hardware. Nevertheless, the same environment definition would transfer to an Isaac-Lab-based training loop with observations, actions, and reward specified as above.

\section{Shared-Dynamics System Architecture}

\subsection{The USD scene as the single parameter source}

The vehicle, the pipeline, the seafloor, and a set of randomizable spherical rock obstacles are described in one Universal Scene Description scene, and the ROV prim carries custom attributes alongside the standard PhysX rigid-body schema: rigid-body mass, an explicit weight and buoyancy pair, six diagonal added-mass coefficients, six linear and six quadratic damping coefficients, a maximum per-thruster force, an eight-thruster (6x8) allocation matrix, and the offset of the centre of buoyancy from the body origin. Every script in the system, whether it trains a reinforcement-learning policy, runs one of the six classical baselines of Section VIII, or flies a controller inside Isaac Sim, parses this same file at start-up. Gravity is disabled at the rigid-body level for the ROV prim, so the full weight-versus-buoyancy restoring force of Section IV-D is supplied explicitly by whichever controller is active, and does not have to be reasoned about separately for the training and deployment models. This architecture's vehicle model uses the real BlueROV2-Heavy mass, added-mass, damping, and eight-thruster allocation matrix sourced from a vehicle-specific reference script (BlueRov\_m\_script.m). All five reinforcement-learning algorithms of Sections VI and VII, and all six classical baselines of Section VIII, are built and trained against the eight-actuator environment.

\subsection{Training and deployment as two consumers of one model}

Figure 1 sets out the two branches that read from the same simulation scene. The right-hand side shows the NVIDIA Isaac Sim 6.0.1 environment for ROV pipeline tracking, comprising a simulated ROV, an L-shaped pipeline, waypoint markers, and surrounding obstacles. The 3D scene provides a physics-based environment for visualising and evaluating ROV navigation and control. In contrast, the training branch uses a NumPy-vectorised Fossen dynamics engine to run 64 environments in parallel with a fixed-step fourth-order Runge–Kutta integrator at a 0.05 s timestep, providing the basis for the five reinforcement-learning trainers in Sections VI and VII and the six classical controllers in Section VIII. The deployment branch loads the same scene in Isaac Sim, attaches a controller to the ROV prim, loads either a trained policy checkpoint or classical control law, generates thruster commands at a rate consistent with the training timestep, and applies the resulting forces to the PhysX rigid body at each physics step, with PhysX performing the integration. The two branches do not exchange state directly; only the trained policy checkpoint or tuned classical-controller parameters are transferred. Both branches implement the governing equation presented in Section IV-A using parameters from the same file, while the numerical differences arising from the integration method and the deployment branch's algebraic treatment of added mass and the Coriolis–centripetal term are discussed in Section IX.

\begin{figure*}[!t]
\renewcommand{\thefigure}{1}
\centering
\includegraphics[width=0.9\textwidth,height=0.8\textheight,keepaspectratio]{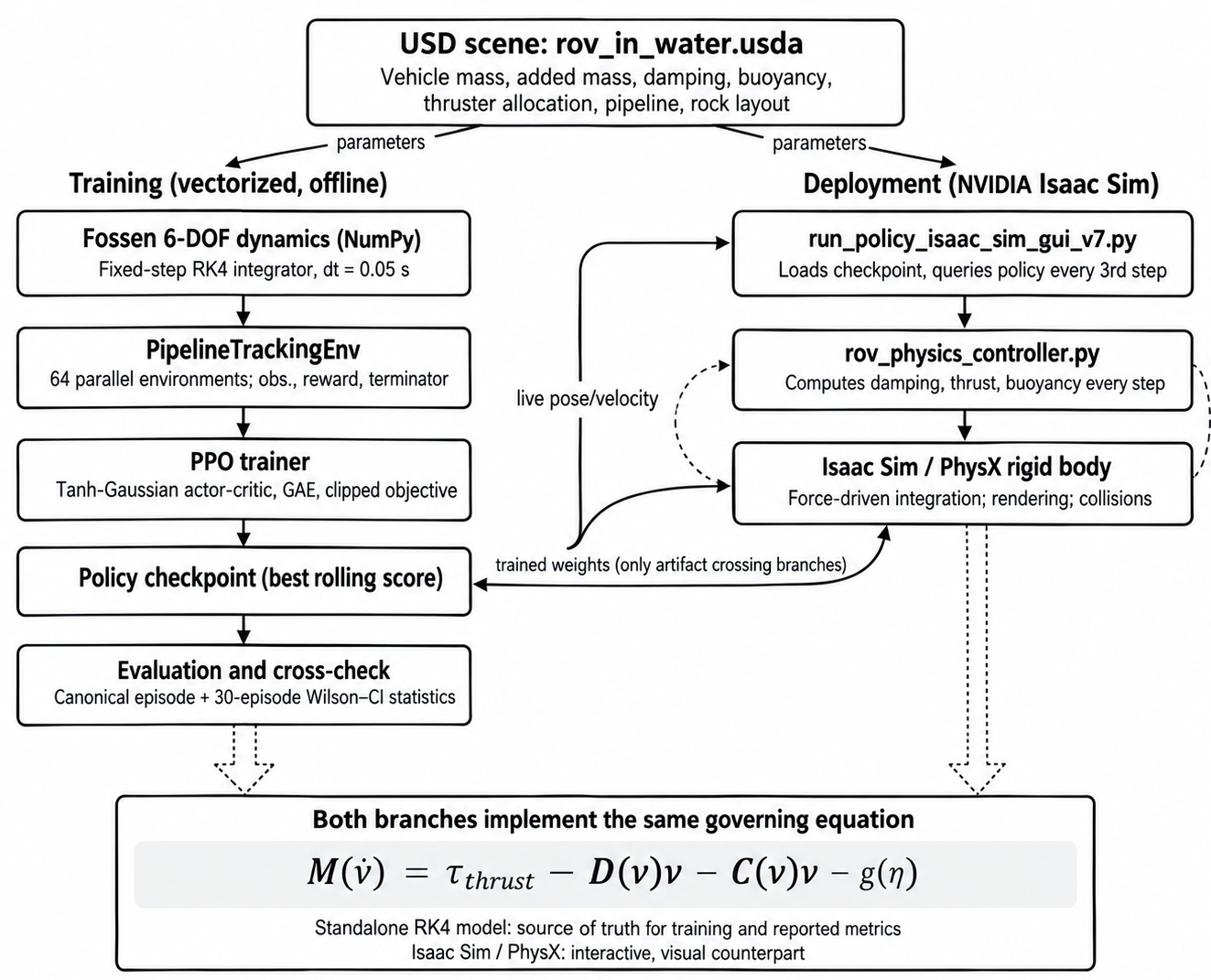}
\caption{System architecture. A single USD scene supplies vehicle and environment parameters to a vectorized training branch, shared across the five compared reinforcement-learning algorithms and the six classical baselines, and an interactive NVIDIA Isaac Sim deployment branch. A trained policy checkpoint or a classical controller's tuned parameters is the only artifact that crosses between them, via a checkpoint-compatibility layer that lets any controller be flown and scored identically.}
\label{fig:1}
\end{figure*}

\section{Fossen Six-Degree-of-Freedom Vehicle Model}

Both branches implement Fossen's marine-craft model in the body-fixed frame \cite{b2}. Let \ensuremath{\nu} = {[}u, v, w, p, q, r{]}\ensuremath{^{T}} collect the body-frame surge, sway, heave, roll-rate, pitch-rate, and yaw-rate velocities, and let \ensuremath{\eta} = {[}x, y, z, \ensuremath{\varphi}, \ensuremath{\theta}, \ensuremath{\psi}{]}\ensuremath{^{T}} collect the world-frame position and Euler-angle attitude. The governing equations are

\begin{equation}
M\dot{\nu} + C(\nu)\nu + D(\nu) + g(\eta)\tag{1} = \tau_{thrust}\label{eq:1}
\end{equation}

\begin{equation}
\dot{\eta} = J(\eta)\nu\tag{2}\label{eq:2}
\end{equation}

where M is the 6\ensuremath{\times}6 mass matrix, C(\ensuremath{\nu}) is the Coriolis-centripetal matrix of Section IV-B, D(\ensuremath{\nu}) collects linear and quadratic damping, g(\ensuremath{\eta}) is the restoring, weight-versus-buoyancy vector, \ensuremath{\tau}\_thrust is the thruster wrench, and J(\ensuremath{\eta}) is the kinematic transform relating body-frame velocity to the rate of change of world-frame pose. The Coriolis-centripetal term is included in this governing equation as the primary dynamics model throughout this paper. Section XI-A reports a controlled ablation, restricted to PPO and TRPO, showing that including the term does not destabilize either algorithm and modestly improves tracking accuracy for both, and Section IX-B describes how the term is applied inside the live Isaac Sim deployment without double-counting PhysX's own intrinsic handling of rigid-body rotation.

\subsection{Mass matrix}

The rigid-body mass matrix M\_RB includes a centre-of-gravity offset r\_g = (r\_gx, r\_gy, r\_gz) coupling surge and sway with pitch and roll. Added mass M\_A is diagonal and applied with the datasheet's sign convention, added-mass coefficients listed as negative numbers, so that

\begin{equation}
M_{A} = - diag\left( m_{added} \right)\tag{3}\label{eq:3}
\end{equation}

\begin{equation}
M = M_{RB} + M_{A}\tag{4}\label{eq:4}
\end{equation}

where M\_A is the diagonal added-mass matrix, m\_added is the vector of six added-mass coefficients read from the scene datasheet, M\_RB is the rigid-body mass matrix, and M is the combined mass matrix used in Section IV's governing equation. See Table I for the parameters used.

\begin{table}[!t]
\renewcommand{\thetable}{I}
\caption{Rigid-body mass and added-mass parameters}
\label{tab:I}
\centering\footnotesize
\setlength{\tabcolsep}{3pt}
\begin{tabularx}{\columnwidth}{@{}>{\raggedright\arraybackslash}X>{\raggedright\arraybackslash}X>{\raggedright\arraybackslash}X@{}}
\hline
{\bfseries Parameter} & {\bfseries Value} & {\bfseries Degree of freedom} \\
\hline
Rigid-body mass, m & 11.5 kg & All \\
Added mass, surge (m\_added,1) & -5.5 kg & Surge, u \\
Added mass, sway (m\_added,2) & -12.7 kg & Sway, v \\
Added mass, heave (m\_added,3) & -14.57 kg & Heave, w \\
Added mass, roll (m\_added,4) & -0.12 kg m\ensuremath{^{2}} & Roll rate, p \\
Added mass, pitch (m\_added,5) & -0.12 kg m\ensuremath{^{2}} & Pitch rate, q \\
Added mass, yaw (m\_added,6) & -0.12 kg m\ensuremath{^{2}} & Yaw rate, r \\
\hline
\end{tabularx}
\par\vspace{3pt}\parbox{\columnwidth}{\scriptsize \emph{Real BlueROV2-Heavy parameters sourced from BlueRov\_m\_script.m (Section III-A). Added-mass coefficients are stored with the reference script's negative sign convention; each contributes the corresponding positive magnitude to the diagonal of M per Section IV-A.}}
\end{table}

\subsection{Coriolis-centripetal matrix}

The Coriolis-centripetal matrix is the sum of a rigid-body term and an added-mass term, C(\ensuremath{\nu}) = C\_RB(\ensuremath{\nu}) + C\_A(\ensuremath{\nu}), each built from the corresponding block of a mass matrix by the standard skew-symmetric cross-product construction used throughout the marine-craft literature \cite{b2}. Writing \ensuremath{\nu}\ensuremath{_{1}} = {[}u,v,w{]}\ensuremath{^{T}} and \ensuremath{\nu}\ensuremath{_{2}} = {[}p,q,r{]}\ensuremath{^{T}} for the linear and angular halves of the body-frame velocity, and M\ensuremath{_{11}}, M\ensuremath{_{12}}, M\ensuremath{_{21}}, M\ensuremath{_{22}} for the corresponding 3\ensuremath{\times}3 blocks of a mass matrix M, the general skew-symmetric Coriolis-centripetal matrix associated with M takes the block form

\begin{equation}
C(\nu) = C_{RB}(\nu) + C_{A}(\nu)\tag{5}\label{eq:5}
\end{equation}

with each 3\ensuremath{\times}3 block of C(\ensuremath{\nu}) built from the cross-product operator S(a)b = a \ensuremath{\times} b applied to the linear-momentum-like vector M\ensuremath{_{11}}\ensuremath{\nu}\ensuremath{_{1}} + M\ensuremath{_{12}}\ensuremath{\nu}\ensuremath{_{2}} and the angular-momentum-like vector M\ensuremath{_{21}}\ensuremath{\nu}\ensuremath{_{1}} + M\ensuremath{_{22}}\ensuremath{\nu}\ensuremath{_{2}}, so that the upper-right, lower-left, and lower-right 3\ensuremath{\times}3 blocks of C(\ensuremath{\nu}) are -S(M\ensuremath{_{11}}\ensuremath{\nu}\ensuremath{_{1}}+M\ensuremath{_{12}}\ensuremath{\nu}\ensuremath{_{2}}), -S(M\ensuremath{_{11}}\ensuremath{\nu}\ensuremath{_{1}}+M\ensuremath{_{12}}\ensuremath{\nu}\ensuremath{_{2}}), and -S(M\ensuremath{_{21}}\ensuremath{\nu}\ensuremath{_{1}}+M\ensuremath{_{22}}\ensuremath{\nu}\ensuremath{_{2}}) respectively, with the upper-left block identically zero \cite{b2}. C\_RB(\ensuremath{\nu}) is built this way from the rigid-body mass blocks M\_RB,11, M\_RB,12, M\_RB,21, M\_RB,22 of Section IV-A, and C\_A(\ensuremath{\nu}) from the diagonal added-mass matrix M\_A of the same section. Because M\_A is diagonal, C\_A(\ensuremath{\nu})\ensuremath{\nu} reduces to a small set of quadratic cross-velocity terms scaled by the differences between along-axis added-mass coefficients: for example, a roll-moment contribution proportional to (m\_added,heave - m\_added,sway)\ensuremath{\cdot}v\ensuremath{\cdot}w, the classic Munk-moment coupling that turns simultaneous sway and heave added-mass reaction into a roll-axis torque. Section IX-B returns to this split between the rigid-body and added-mass components when describing how the term is applied inside the live Isaac Sim deployment.

\subsection{Damping}

Damping combines a linear and a quadratic term per degree of freedom,

\begin{equation}
D(\nu) = D_{lin}\nu + D_{quad}\nu|\nu|\tag{6}\label{eq:6}
\end{equation}

where D(\ensuremath{\nu}) is the hydrodynamic damping force and moment vector, D\_lin and D\_quad are diagonal matrices of linear and quadratic damping coefficients, and \ensuremath{\nu}\textbar \ensuremath{\nu}\textbar{} denotes each component of \ensuremath{\nu} multiplied by its own absolute value. See Table II for the parameters used.

\begin{table}[!t]
\renewcommand{\thetable}{II}
\caption{Linear and quadratic damping coefficients}
\label{tab:II}
\centering\footnotesize
\setlength{\tabcolsep}{3pt}
\begin{tabularx}{\columnwidth}{@{}>{\raggedright\arraybackslash}X>{\raggedright\arraybackslash}X>{\raggedright\arraybackslash}X>{\raggedright\arraybackslash}X@{}}
\hline
{\bfseries Degree of freedom} & {\bfseries D\_lin} & {\bfseries D\_quad} & {\bfseries Units} \\
\hline
Surge, u & -4.03 & -18.18 & N per m/s; N per (m/s)\ensuremath{^{2}} \\
Sway, v & -6.22 & -21.66 & N per m/s; N per (m/s)\ensuremath{^{2}} \\
Heave, w & -5.18 & -36.99 & N per m/s; N per (m/s)\ensuremath{^{2}} \\
Roll rate, p & -0.07 & -1.55 & N\ensuremath{\cdot}m per rad/s; N\ensuremath{\cdot}m per (rad/s)\ensuremath{^{2}} \\
Pitch rate, q & -0.07 & -1.55 & N\ensuremath{\cdot}m per rad/s; N\ensuremath{\cdot}m per (rad/s)\ensuremath{^{2}} \\
Yaw rate, r & -0.07 & -1.55 & N\ensuremath{\cdot}m per rad/s; N\ensuremath{\cdot}m per (rad/s)\ensuremath{^{2}} \\
\hline
\end{tabularx}
\par\vspace{3pt}\parbox{\columnwidth}{\scriptsize \emph{Diagonal entries of D\_lin and D\_quad in the governing damping term D(\ensuremath{\nu}) = D\_lin\ensuremath{\nu} + D\_quad\ensuremath{\nu}\textbar \ensuremath{\nu}\textbar{} (Section IV-C). Signed negative per the datasheet convention, so the term always opposes motion.}}
\end{table}

\subsection{Restoring forces}

With a z-up coordinate system and net weight-minus-buoyancy W\_net = W - B, the translational components of the restoring vector are

\begin{equation}
g_{x} = - W_{net}\sin(\theta)\tag{7}\label{eq:7}
\end{equation}

\begin{equation}
g_{y} = W_{net}\cos(\theta)\sin(\varphi)\tag{8}\label{eq:8}
\end{equation}

\begin{equation}
g_{z} = W_{net}\cos(\theta)\cos(\varphi)\tag{9}\label{eq:9}
\end{equation}

where g\_x, g\_y, g\_z are the body-frame components of the restoring vector g(\ensuremath{\eta}), W\_net = W - B is the net weight-minus-buoyancy force, and \ensuremath{\varphi}, \ensuremath{\theta} are the vehicle's roll and pitch Euler angles. See Table III for the parameters used.

\begin{table}[!t]
\renewcommand{\thetable}{III}
\caption{Weight, buoyancy, and net restoring force}
\label{tab:III}
\centering\footnotesize
\setlength{\tabcolsep}{3pt}
\begin{tabularx}{\columnwidth}{@{}>{\raggedright\arraybackslash}X>{\raggedright\arraybackslash}X@{}}
\hline
{\bfseries Parameter} & {\bfseries Value} \\
\hline
Weight, W & 112.8 N \\
Buoyancy, B & 114.8 N \\
Net weight-minus-buoyancy, W\_net = W - B & -2.0 N (positively buoyant) \\
\hline
\end{tabularx}
\par\vspace{3pt}\parbox{\columnwidth}{\scriptsize \emph{Values entering the restoring-force terms g\_x, g\_y, g\_z of Section IV-D. A negative W\_net confirms the vehicle is positively buoyant by design, consistent with a safe passive ascent if power is lost.}}
\end{table}

\subsection{Thruster allocation}

Eight normalized thruster commands u\_i in the range -1 to 1 map to per-thruster forces and then to a body-frame wrench through a 6\ensuremath{\times}8 allocation matrix A read from the scene:

\begin{equation}
f_{i} = \operatorname{clip}\left( u_{i}, - 1,1 \right) \cdot T_{\max},\ i = 1,...,8\tag{10}\label{eq:10}
\end{equation}

\begin{equation}
\tau_{thrust} = Af\tag{11}\label{eq:11}
\end{equation}

where f\_i is the force commanded by thruster i, u\_i is the normalized command for that thruster in {[}-1,1{]}, T\_max is the maximum per-thruster force, f collects all eight thruster forces, A is the thruster allocation matrix read from the scene, and \ensuremath{\tau}\_thrust is the resulting body-frame thrust wrench.

Maximum per-thruster force T\_max is 40 N. This architecture uses the real BlueROV2-Heavy eight-thruster arrangement (see Figure 4). Four horizontal thrusters (T1-T4) in an X pattern angled at 45 degrees at the vehicle's corners, and four vertical thrusters (T5-T8), one per hull corner. Figure 3 reproduces the manufacturer's own thruster force/torque diagram, and Fig. 4 shows this project's own T1-T8 layout as generated from the scene-construction script's geometry; the two use different thruster indices for the same physical arrangement.

\begin{figure}[!t]
\renewcommand{\thefigure}{3}
\centering
\includegraphics[width=\columnwidth,height=0.8\textheight,keepaspectratio]{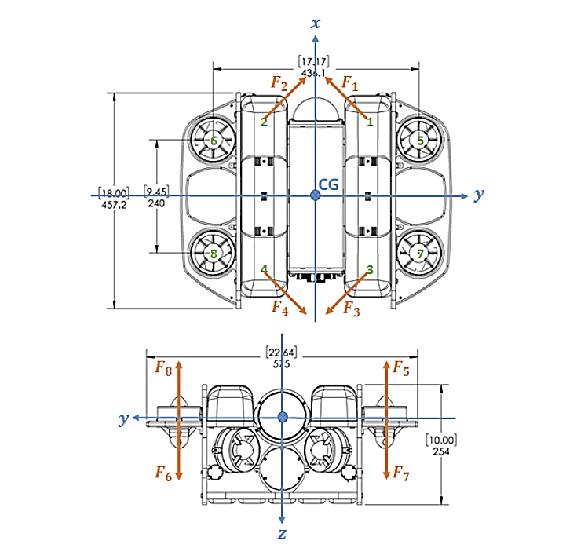}
\caption{BlueROV2-Heavy thruster force/torque diagram. See URL for more details: https://bluerobotics.com/introducing-bluerov2-heavy/}
\label{fig:3}
\end{figure}

\begin{figure*}[!t]
\renewcommand{\thefigure}{4}
\centering
\includegraphics[width=0.9\textwidth,height=0.8\textheight,keepaspectratio]{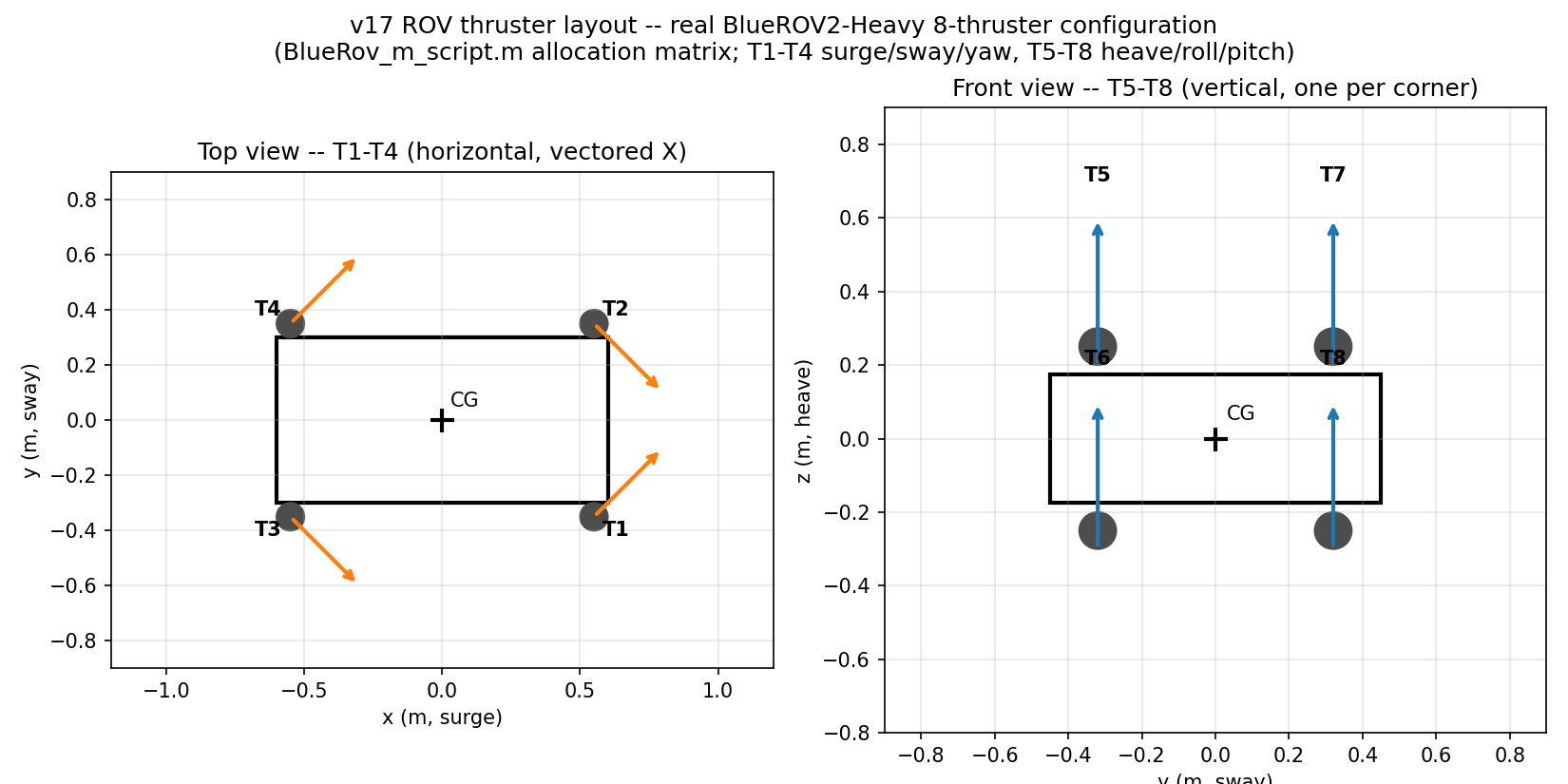}
\caption{This project's thruster layout (indices T1-T8), drawn from the scene-construction script's actual geometry: T1-T4 horizontal in an X pattern at 45 degrees, T5-T8 vertical, one per hull corner.}
\label{fig:4}
\end{figure*}

\begin{table}[!t]
\renewcommand{\thetable}{IV}
\caption{Thruster allocation parameters}
\label{tab:IV}
\centering\footnotesize
\setlength{\tabcolsep}{3pt}
\begin{tabularx}{\columnwidth}{@{}>{\raggedright\arraybackslash}X>{\raggedright\arraybackslash}X@{}}
\hline
{\bfseries Parameter} & {\bfseries Value} \\
\hline
Number of thrusters & 8 (T1-T4 horizontal, T5-T8 vertical) \\
Maximum per-thruster force, T\_max & 40 N \\
Normalized command range, u\_i & {[}-1, 1{]} per thruster \\
Allocation matrix, A & 6\ensuremath{\times}8, read from the USD scene (Table IV-B) \\
\hline
\end{tabularx}
\end{table}

\begin{table*}[!t]
\renewcommand{\thetable}{IV-B}
\caption{Thruster allocation matrix A (per-thruster contribution to surge u, sway v, heave w, roll p, pitch q, yaw r)}
\label{tab:IV-B}
\centering\footnotesize
\setlength{\tabcolsep}{3pt}
\begin{tabularx}{\textwidth}{@{}>{\raggedright\arraybackslash}X>{\raggedright\arraybackslash}X>{\raggedright\arraybackslash}X>{\raggedright\arraybackslash}X>{\raggedright\arraybackslash}X>{\raggedright\arraybackslash}X>{\raggedright\arraybackslash}X@{}}
\hline
{\bfseries Thruster} & {\bfseries u} & {\bfseries v} & {\bfseries w} & {\bfseries p} & {\bfseries q} & {\bfseries r} \\
\hline
T1 & 0.707 & -0.707 & 0 & 0.060 & 0.060 & -0.189 \\
T2 & 0.707 & 0.707 & 0 & -0.060 & 0.060 & 0.189 \\
T3 & -0.707 & -0.707 & 0 & 0.060 & -0.060 & 0.189 \\
T4 & -0.707 & 0.707 & 0 & -0.060 & -0.060 & -0.189 \\
T5 & 0 & 0 & -1 & -0.218 & 0.120 & 0 \\
T6 & 0 & 0 & 1 & -0.218 & -0.120 & 0 \\
T7 & 0 & 0 & 1 & 0.218 & 0.120 & 0 \\
T8 & 0 & 0 & -1 & 0.218 & -0.120 & 0 \\
\hline
\end{tabularx}
\par\vspace{3pt}\parbox{\textwidth}{\scriptsize \emph{Numeric entries of A (Section IV-E), read directly from the USD scene. T1-T4 are the horizontal, 45-degree X-pattern thrusters and contribute no heave; T5-T8 are the vertical thrusters and contribute no surge or sway. Units are per unit normalized command, so each entry multiplies the corresponding f\_i of Section IV-E before summation into \ensuremath{\tau}\_thrust.}}
\end{table*}

\subsection{Kinematics and time integration}

Position and Euler-angle rates use the body-to-world rotation matrix R(\ensuremath{\varphi}, \ensuremath{\theta}, \ensuremath{\psi}) and the Euler-rate transform T(\ensuremath{\varphi}, \ensuremath{\theta}):

\begin{equation}
\frac{d\ pos}{dt} = R(\varphi,\theta,\psi){\lbrack u,v,w\rbrack}^{T}\tag{12}\label{eq:12}
\end{equation}

\begin{equation}
\frac{d\ \eta_{rot}}{dt} = T(\varphi,\theta){\lbrack p,q,r\rbrack}^{T}\tag{13}\label{eq:13}
\end{equation}

where pos = {[}x,y,z{]}\ensuremath{^{T}} is the world-frame position, R(\ensuremath{\varphi},\ensuremath{\theta},\ensuremath{\psi}) is the body-to-world rotation matrix, \ensuremath{\eta}\_rot = {[}\ensuremath{\varphi},\ensuremath{\theta},\ensuremath{\psi}{]}\ensuremath{^{T}} is the Euler-angle attitude, and T(\ensuremath{\varphi},\ensuremath{\theta}) is the Euler-rate transform relating body-frame angular velocity {[}p,q,r{]}\ensuremath{^{T}} to the rate of change of attitude.

T is singular at \ensuremath{\theta} = \ensuremath{\pm}90 degrees, so \ensuremath{\theta} is clamped to approximately \ensuremath{\pm}85 degrees before T is evaluated. The standalone training and evaluation model integrates the full state with a fixed-step fourth-order Runge-Kutta scheme at dt = 0.05 s (20 Hz):

\begin{equation}
k_{1} = f\left( x_{t},u \right)\tag{14}\label{eq:14}
\end{equation}

\begin{equation}
k_{2} = f\left( x_{t} + \frac{dt}{2}k_{1},u \right)\tag{15}\label{eq:15}
\end{equation}

\begin{equation}
k_{3} = f\left( x_{t} + \frac{dt}{2}k_{2},u \right)\tag{16}\label{eq:16}
\end{equation}

\begin{equation}
k_{4} = f\left( x_{t} + dt\ k_{3},u \right)\tag{17}\label{eq:17}
\end{equation}

\begin{equation}
x_{t + 1} = x_{t} + \frac{dt}{6}\left( k_{1} + 2k_{2} + 2k_{3} + k_{4} \right)\tag{18}\label{eq:18}
\end{equation}

where x\_t = {[}\ensuremath{\nu},\ensuremath{\eta}{]} is the full state at step t, u is the thruster command held constant over the step, f(x,u) evaluates the right-hand side of Section IV's governing equations, dt = 0.05 s is the fixed integration timestep, and k1 through k4 are the four Runge-Kutta stage derivatives.

Section IX describes how this integrator is reconciled with Isaac Sim, which integrates the same forces via PhysX rather than a second Runge-Kutta scheme.

\section{Reinforcement Learning Environment}

The Fossen model of Section IV is wrapped in a vectorized, Gym-style environment, PipelineTrackingEnv, running 64 independent vehicle instances in NumPy. The environment, reward, and safety limits are identical across all five reinforcement-learning algorithms of Sections VI and VII, and the same guidance signals feed the six classical baselines of Section VIII.

\subsection{Episode length and control rate}

Episodes run for up to 1,040 steps of 0.05 s each (52 s), a budget set from the nominal transit time of the roughly 10 m L-shaped pipeline route with headroom for manoeuvring around its one corner and the randomized rocks. The policy acts at 20 Hz; inside the live Isaac Sim deployment of Section IX, physics runs at 60 Hz and the controller is queried once every third step.

\subsection{Observation}

The observation vector has 25 dimensions, summarized in Table V: position (3), the sine and cosine of the three Euler angles (6), body-frame velocity (6), along-track progress and remaining distance (2), horizontal and vertical cross-track error (2), three-dimensional distance to the pipeline centerline (1), the cosine similarity between the vehicle's forward axis and the local pipeline tangent (1), and four obstacle-relative terms describing the nearest rock's distance, bearing, and elevation (4).

\begin{table}[!t]
\renewcommand{\thetable}{V}
\caption{Observation vector (25 dimensions)}
\label{tab:V}
\centering\footnotesize
\setlength{\tabcolsep}{3pt}
\begin{tabularx}{\columnwidth}{@{}>{\raggedright\arraybackslash}X>{\raggedright\arraybackslash}X>{\raggedright\arraybackslash}X@{}}
\hline
{\bfseries Group} & {\bfseries Dims} & {\bfseries Contents} \\
\hline
Position & 3 & x, y, z \\
Orientation & 6 & sin and cos of \ensuremath{\varphi}, \ensuremath{\theta}, \ensuremath{\psi} \\
Body velocity & 6 & \ensuremath{\nu} = {[}u, v, w, p, q, r{]} \\
Route progress & 2 & Along-track distance s; distance to goal \\
Cross-track error & 2 & Horizontal cross\_h; vertical cross\_v \\
Pipe standoff & 1 & 3D distance to pipeline centerline \\
Heading & 1 & Cosine similarity, forward axis vs.~local tangent \\
Obstacle & 4 & Nearest-rock distance; sin/cos bearing; elevation \\
\hline
\end{tabularx}
\end{table}

\subsection{Reward}

Writing \ensuremath{\Delta}s for the along-track progress made in one step, e for the three-dimensional distance from the vehicle to the pipeline centerline, cross\_h for the horizontal component of that error, and a for the heading-alignment term in {[}-1, 1{]}, the reward combines a progress term, a quadratic three-dimensional standoff-tracking term and its rate of change, a quadratic horizontal-tracking term, a heading term, an energy penalty, and terminal collision and completion terms, using the gains in Table VI. The three-dimensional tracking term is a quadratic attractor centred on the 1.2 m standoff target; the horizontal-tracking term pulls the horizontal component of standoff error toward a tighter 0.2 m target, leaving the vertical component to absorb the remainder of the 1.2 m overall standoff and the 1.0 m hard safety floor.

\begin{table}[!t]
\renewcommand{\thetable}{VI}
\caption{Reward weights and safety limits (identical for all five reinforcement-learning algorithms)}
\label{tab:VI}
\centering\footnotesize
\setlength{\tabcolsep}{3pt}
\begin{tabularx}{\columnwidth}{@{}>{\raggedright\arraybackslash}X>{\raggedright\arraybackslash}X>{\raggedright\arraybackslash}X@{}}
\hline
{\bfseries Parameter} & {\bfseries Value} & {\bfseries Meaning} \\
\hline
k\_progress & 48 & Reward for along-track progress \\
k\_track, k\_track\_rate & 20, 0.5 & Penalty for 3D standoff error (1.2 m target) and its rate \\
k\_cross\_h, cross\_h\_target & 15, 0.2 m & Horizontal cross-track penalty and target \\
k\_heading & 1.5 & Reward for facing along the pipeline \\
k\_energy & 1.0 & Thruster-effort penalty \\
completion\_bonus & 100 & Bonus for reaching the end \\
collision\_penalty & 50 & Penalty for a collision \\
standoff\_target (hard floor) & 1.0 m & Minimum allowed gap to the pipe \\
waypoint\_radius & 0.30 m & Distance at which the route end counts as reached \\
\hline
\end{tabularx}
\end{table}

\subsection{Termination and domain randomization}

Each episode terminates under one of six conditions, four of which count as failures. A collision is registered when the vehicle's distance to any rock surface drops below its collision radius, or when the vehicle reaches the seafloor. A pipe-gap violation occurs once the distance to the centerline falls below the 1.0 m hard safety floor. A flip is declared when the absolute roll or pitch angle exceeds 75 degrees, and the episode also ends if the vehicle leaves the operating bounds. Success is recorded when the vehicle comes within the 0.30 m waypoint radius of the route's final point. Any episode that reaches the maximum length without meeting one of these conditions ends in timeout. At every reset, the horizontal position and radius of each rock are jittered within a narrow band, so the policy is never trained against an identical obstacle layout twice.

\subsection{A corner-cutting projection exploit}

The pipeline geometry includes a single 90-degree turn, and along-track progress is credited by projecting the vehicle's position onto the nearest point of a piecewise-linear centerline, one segment at a time. With that projection computed independently per segment, a diagonal path can project favorably onto the far segment of the corner without the vehicle approaching the true corner vertex, collecting progress reward for a route it never flew; one diagnosed episode reached 9.31 m of a 10.00 m route while never coming closer than 1.86 m to the vertex. A per-environment projection gate, requiring the vehicle to close within a fixed radius of a segment's end vertex before projection onto the next segment is permitted, was verified against a synthetic corner-cutting path (rejected, growing tracked error) and a genuine cornering path (unlocked and tracked correctly). This projection exploit addresses how progress reward is credited by the guidance geometry itself, independently of any particular algorithm's or controller's own training or tuning dynamics, and the fix, together with the look-ahead heading point of Section VIII's closing paragraph, benefits every reinforcement-learning and classical controller compared in this paper equally.

\section{On-Policy Optimization: Proximal Policy Optimization}

PPO serves as the on-policy baseline against which SAC, TD3, DDPG, and TRPO are compared, and, together with TRPO and feedback linearization, is one of the strongest performers once the classical baselines of Section VIII are added to the comparison (Section XI).

\subsection{Tanh-squashed Gaussian policy}

The actor network outputs a raw, pre-tanh mean; log-probabilities and PPO's importance-sampling ratios are computed on this raw Gaussian, while the bounded thruster command sent to the environment is the hyperbolic tangent of a sample from it:

\begin{equation}
a_{raw}\mathcal{\sim N(}\mu(s),\sigma)\tag{19}\label{eq:19}
\end{equation}

\begin{equation}
a = tanh\left( a_{raw} \right) \in \lbrack - 1,1\rbrack\tag{20}\label{eq:20}
\end{equation}

\begin{equation}
\log\ \pi\left( a|s \right) = log\ \mathcal{N}(a_{raw}) - \sum{\log\left( 1 - \tanh^{2}\left( a_{raw} \right) + \varepsilon \right)}\tag{21}\label{eq:21}
\end{equation}

where a\_raw is a sample from the actor's raw Gaussian with mean \ensuremath{\mu}(s) and standard deviation \ensuremath{\sigma}, a is the bounded thruster command applied to the environment, and the summation subtracts the tanh-squashing Jacobian correction (with a small constant \ensuremath{\varepsilon} for numerical stability) from the raw Gaussian log-probability to give the log-probability of the squashed action.

Actions are squashed smoothly with tanh. Hard-clipping an unbounded Gaussian sets the policy gradient to zero for any sample beyond the action bound, while the tanh mapping keeps it nonzero across the full range. All five reinforcement-learning algorithms compared in this paper apply the same tanh nonlinearity, so each shares a common action interface with the environment and with the deployment controller of Section IX.

\subsection{Generalized advantage estimation}

Advantages are computed with generalized advantage estimation \cite{b13},

\begin{equation}
\delta_{t} = r_{t} + \gamma V\left( s_{t + 1} \right)\left( 1 - d_{t} \right) - V\left( s_{t} \right)\tag{22}\label{eq:22}
\end{equation}

\begin{equation}
A_{t} = \sum_{l = 0}^{T}{(\gamma\lambda)^{l}\delta_{t + l}}\tag{23}\label{eq:23}
\end{equation}

\begin{equation}
{Return}_{t} = A_{t} + V\left( s_{t} \right)\tag{24}\label{eq:24}
\end{equation}

where \ensuremath{\delta}\_t is the temporal-difference residual at step t, r\_t is the reward, \ensuremath{\gamma} is the discount factor, V(s) is the critic's value estimate, d\_t is a done flag, \ensuremath{\lambda} is the GAE trace-decay parameter, A\_t is the estimated advantage, and Return\_t is the value-function regression target.

\subsection{Clipped surrogate objective}

The policy is updated with the clipped surrogate objective of proximal policy optimization \cite{b12}:

\begin{equation}
r_{t}(\theta) = \operatorname{exp}\left( \log\ \pi_{\theta}\left( a_{t}|s_{t} \right) - log\ \pi_{\theta old}\left( a_{t}|s_{t} \right) \right)\tag{25}\label{eq:25}
\end{equation}

\begin{equation}
\adjustbox{max width=\columnwidth}{$\displaystyle L^{CLIP}\mathbb{= E}\lbrack \operatorname{min}\left( r_{t}(\theta)A_{t},\ \operatorname{clip}\left( r_{t}(\theta),1 - \varepsilon,1 + \varepsilon \right)A_{t} \right)\rbrack$}\tag{26}\label{eq:26}
\end{equation}

\begin{equation}
Loss = - L^{CLIP} + c_{1}L^{VF} - c_{2}H\lbrack\pi\rbrack\tag{27}\label{eq:27}
\end{equation}

where r\_t(\ensuremath{\theta}) is the importance-sampling ratio between the current and previous policy, A\_t is the advantage of Section VI-B, \ensuremath{\varepsilon} is the clip range, L\^{}CLIP is the clipped surrogate objective, L\^{}VF is the value-function loss, H{[}\ensuremath{\pi}{]} is the policy entropy, and c1, c2 weight the value and entropy terms in the total loss.

\subsection{Training configuration and checkpoint selection}

Table VII lists the hyperparameters for the final PPO training run. PPO and TRPO both follow a curriculum in which the agent trains on a short, one-segment scene for the first 35 percent of the training budget and then switches to the full route. Late in training, PPO's entropy bonus can pull the policy away from a deterministic solution that already performs well. To guard against this, the trainer keeps a rolling score over recently finished episodes and, once enough episodes have completed, saves the checkpoint with the highest score:

\begin{equation}
score = p_{success} - 0.5\ p_{flip} - 0.5\ p_{gap}\tag{28}\label{eq:28}
\end{equation}

where p\_success, p\_flip, and p\_gap are the rolling success, flip, and pipe-gap-violation rates over recently finished episodes, and the checkpoint with the highest score is the one saved.

This criterion is reused unmodified for SAC, TD3, DDPG, and TRPO via the compatibility layer of Section VII-F.

\begin{table}[!t]
\renewcommand{\thetable}{VII}
\caption{PPO training hyperparameters (final accepted run)}
\label{tab:VII}
\centering\footnotesize
\setlength{\tabcolsep}{3pt}
\begin{tabularx}{\columnwidth}{@{}>{\raggedright\arraybackslash}X>{\raggedright\arraybackslash}X@{}}
\hline
{\bfseries Hyperparameter} & {\bfseries Value} \\
\hline
Total timesteps (final run) & 20,004,864 \\
Parallel environments & 64 \\
Rollout length & 256 steps \\
Curriculum fraction (short scene first) & 0.35 \\
Learning rate & 3\ensuremath{\times}10\ensuremath{^{-4}}, Adam \\
PPO epochs per update & 10 \\
Minibatch size & 4,096 \\
Discount, \ensuremath{\gamma} & 0.99 \\
GAE, \ensuremath{\lambda} & 0.95 \\
Clip ratio, \ensuremath{\varepsilon} & 0.2 \\
Value-function coefficient, c\ensuremath{_{1}} & 0.5 \\
Entropy coefficient, c\ensuremath{_{2}} & 0.01 \\
Max gradient norm & 0.5 \\
Network & MLP, 2 hidden layers of 256 units, tanh \\
\hline
\end{tabularx}
\end{table}

\section{Off-Policy and Trust-Region Alternatives}

SAC, TD3, DDPG, and TRPO are each implemented as a standalone trainer. Every trainer imports the unchanged PipelineTrackingEnv of Section V, and its policies are scored by the unchanged evaluation functions of Section X through the compatibility layer of Section VII-F. All networks match PPO's own multilayer perceptron, with two hidden layers of 256 units and tanh activations, which holds architectural capacity constant across the comparison.

\subsection{Design considerations for replay-buffer training}

Domain-randomized rock positions work with a replay buffer because the distance and bearing to the nearest rock are fully observed. The curriculum schedule, which trains first on a short scene and then on the full-length scene, causes a problem for off-policy methods. Total pipeline length differs between the two phases and is not part of the observation. The same observed along-track distance can therefore correspond to a different true reward depending on which phase generated it, and a persistent buffer holding transitions from both phases would alias them together. For this reason, the curriculum is disabled for SAC, TD3, and DDPG. TRPO is on-policy, so it retains the curriculum. To manage the sensitivity of SAC, TD3, and DDPG to reward magnitude, a global multiplier of reward\_scale = 0.1 multiplier is applied only to the critic's temporal-difference target, leaving the reward's own gains unchanged.

\subsection{Soft actor-critic (SAC)}

SAC augments the standard return with an entropy bonus weighted by a temperature \ensuremath{\alpha}, favouring stochastic policies that remain as random as possible while still maximizing return \cite{b14}:

\begin{equation}
\adjustbox{max width=\columnwidth}{$\displaystyle J(\pi) = \sum_{t = 0}^{T}{\mathbb{E}_{(s\ensuremath{_{t}},a\ensuremath{_{t}}) \sim \rho\pi}\lbrack r(s\ensuremath{_{t}},a\ensuremath{_{t}}) + \alpha\ H\left( \pi\left( \cdot |s\ensuremath{_{t}} \right) \right)\rbrack}$}\tag{29}\label{eq:29}
\end{equation}

\begin{equation}
\adjustbox{max width=\columnwidth}{$\displaystyle y = r + \gamma\left( \min_{i = 1,2}\ Q_{\overline{\theta}\ensuremath{_{i}}}\left( s',\widetilde{a}' \right) - \alpha\ \log\ \pi_{\varphi}\left( \widetilde{a}'|s' \right) \right),\ \widetilde{a}' \sim \pi_{\varphi}\left( \cdot |s' \right)$}\tag{30}\label{eq:30}
\end{equation}

\begin{equation}
\adjustbox{max width=\columnwidth}{$\displaystyle J_{\pi}(\varphi) = \mathbb{E}_{s,\varepsilon}\lbrack\alpha\ \log\ \pi_{\varphi}\left( f_{\varphi}(\varepsilon;s)|s \right) - Q_{\theta}\left( s,f_{\varphi}(\varepsilon;s) \right)\rbrack$}\tag{31}\label{eq:31}
\end{equation}

\begin{equation}
\adjustbox{max width=\columnwidth}{$\displaystyle J(\alpha) = \mathbb{E}_{a\ensuremath{_{t}} \sim \pi \ensuremath{_{t}}}\lbrack - \alpha\ \log\ \pi_{t}\left( a\ensuremath{_{t}}|s\ensuremath{_{t}} \right) - \alpha\ \overline{H}\rbrack$}\tag{32}\label{eq:32}
\end{equation}

where \ensuremath{\alpha} is the temperature coefficient weighting the entropy bonus H, Q\_\ensuremath{\theta}1 and Q\_\ensuremath{\theta}2 are the twin critics, \ensuremath{\bar{\ensuremath{\theta}}}1 and \ensuremath{\bar{\ensuremath{\theta}}}2 their delayed target networks, y is the soft Bellman target, f\_\ensuremath{\varphi}(\ensuremath{\varepsilon};s) is the reparameterized actor sample, and \ensuremath{\bar{H}} is the target entropy.

Twin critics Q\_\ensuremath{\theta}1, Q\_\ensuremath{\theta}2 are regressed toward a soft Bellman target built from the minimum of two delayed target critics minus the entropy term of the sampled next action. The tanh-Gaussian actor is updated by the reparameterization trick, sampling an action through a differentiable function f\_\ensuremath{\varphi}(\ensuremath{\varepsilon};s) of an independent noise variable \ensuremath{\varepsilon}, minimizing the divergence between the policy and the Boltzmann distribution induced by Q\_\ensuremath{\theta}. The temperature \ensuremath{\alpha} is tuned automatically against a target entropy \ensuremath{\bar{H}} rather than fixed by hand. In the released configuration, \ensuremath{\gamma} = 0.95 (reduced from a default of 0.99 to shorten SAC's effective planning horizon relative to the route's terminal completion bonus), target entropy is -1.8, and log \ensuremath{\alpha} is floored at ln 0.018 so the temperature can decay but not vanish entirely.

\subsection{Twin-delayed DDPG (TD3)}

TD3 retains a deterministic actor but addresses DDPG's single-critic overestimation bias with three changes \cite{b15}:

\begin{equation}
\widetilde{a}' = \pi_{\varphi'}(s') + \varepsilon,\ \varepsilon \sim \operatorname{clip}\left( \mathcal{N}(0,\sigma), - c,c \right)\tag{33}\label{eq:33}
\end{equation}

\begin{equation}
y = r + \gamma\ \min_{i = 1,2}\ Q_{\theta'\ensuremath{_{i}}}\left( s',\widetilde{a}' \right)\tag{34}\label{eq:34}
\end{equation}

\begin{equation}
L(\theta_{i}\mathbb{) = E\lbrack}\left( Q_{\theta i}(s,a) - y \right)^{2}\rbrack,\ i = 1,2\tag{35}\label{eq:35}
\end{equation}

\begin{equation}
\nabla_{\varphi}J(\varphi)\mathbb{= E\lbrack}\nabla_{a}Q_{\theta 1}(s,a)|_{a = \pi\varphi(s)}\ \nabla_{\varphi}\pi_{\varphi}(s)\rbrack\tag{36}\label{eq:36}
\end{equation}

where \ensuremath{\pi}\_\ensuremath{\varphi}' is the target actor, \ensuremath{\sigma} is the target-policy smoothing noise, c is the noise clip, Q\_\ensuremath{\theta}1 and Q\_\ensuremath{\theta}2 are the twin critics, \ensuremath{\theta}'\ensuremath{_{1}} and \ensuremath{\theta}'\ensuremath{_{2}} their target networks, y is the clipped double-Q target, and the actor is updated using only the first critic Q\_\ensuremath{\theta}1.

Target-policy smoothing adds clipped Gaussian noise to the target action before the bootstrap target is formed. Clipped double-Q learning then takes the minimum over two target critics, and both critics are regressed toward this shared target. The deterministic actor is updated once for every two critic updates, following the gradient of the first critic alone with respect to the action. The released configuration uses \ensuremath{\sigma} = 0.20 exploration noise, a policy-noise clip of c = 0.5, and a delay of two critic updates per actor update.

\subsection{Deep deterministic policy gradient (DDPG)}

DDPG extends the deterministic policy gradient theorem to an actor-critic architecture with a replay buffer and target networks \cite{b16}:

\begin{equation}
\nabla_{\varphi}J(\varphi)\mathbb{\approx E\lbrack}\nabla_{a}Q_{\theta}(s,a)|_{a = \mu\varphi(s)}\ \nabla_{\varphi}\mu_{\varphi}(s)\rbrack\tag{37}\label{eq:37}
\end{equation}

\begin{equation}
y = r + \gamma Q_{\theta'}\left( s',\mu_{\varphi'}(s') \right)\tag{38}\label{eq:38}
\end{equation}

\begin{equation}
L(\theta)\mathbb{= E\lbrack}\left( Q_{\theta}(s,a) - y \right)^{2}\rbrack\tag{39}\label{eq:39}
\end{equation}

\begin{equation}
\theta' \leftarrow \tau\theta + (1 - \tau)\theta',\ \varphi' \leftarrow \tau\varphi + (1 - \tau)\varphi'\tag{40}\label{eq:40}
\end{equation}

where \ensuremath{\mu}\_\ensuremath{\varphi} is the deterministic actor, Q\_\ensuremath{\theta} is the single critic, \ensuremath{\mu}\_\ensuremath{\varphi}' and Q\_\ensuremath{\theta}' are their target networks, y is the bootstrap target, and \ensuremath{\tau} is the Polyak-averaging rate.

The actor's update follows the gradient of a single critic with respect to the action, evaluated at the action the deterministic policy would produce. The critic itself is trained toward a bootstrapped target computed from the target networks \ensuremath{\mu}\_\ensuremath{\varphi}' and Q\_\ensuremath{\theta}', and both target networks track their online counterparts through Polyak averaging with rate \ensuremath{\tau} = 0.005. During rollout, exploration comes from independent Gaussian noise added to the raw, pre-tanh action, with \ensuremath{\sigma} = 0.10 in the released configuration.

\subsection{Trust region policy optimization (TRPO)}

TRPO uses PPO's own actor-critic network, replacing the clipped surrogate update with a Kullback-Leibler-constrained natural-gradient step \cite{b17}:

\begin{equation}
\adjustbox{max width=\columnwidth}{$\displaystyle \max_{\theta}\ \mathbb{E}_{s,a \sim \pi\theta old}\lbrack\frac{\pi_{\theta}\left( a|s \right)}{\pi_{\theta old}\left( a|s \right)}\ A^{\theta old}(s,a)\rbrack$}\tag{41}\label{eq:41}
\end{equation}

\begin{equation}
\adjustbox{max width=\columnwidth}{$\displaystyle s.t.\ \mathbb{E}_{s \sim \pi\theta old}\lbrack D_{KL}\left( \pi_{\theta old}\left( \cdot |s \right) \parallel \pi_{\theta}\left( \cdot |s \right) \right)\rbrack \leq \delta$}\tag{42}\label{eq:42}
\end{equation}

\begin{equation}
\theta_{k + 1} = \theta_{k} + \alpha^{j}\ \sqrt{\frac{2\delta}{g^{T}\ H^{- 1}g}}\ H^{- 1}g\tag{43}\label{eq:43}
\end{equation}

where \ensuremath{\pi}\_\ensuremath{\theta}old is the policy before the update, A\^{}\ensuremath{\theta}old is its advantage, D\_KL is the mean KL divergence bounded by \ensuremath{\delta}, g is the policy gradient, H is the Fisher-information matrix, H\ensuremath{^{-1}}g is the natural-gradient direction, and \ensuremath{\alpha}\^{}j is the backtracking line-search step size at iteration j.

Each update maximizes the expected importance-weighted advantage, subject to a bound \ensuremath{\delta} on the mean KL divergence between the old and new policies. The constrained step is approximated in three stages. Conjugate gradient first solves for the natural-gradient direction H\ensuremath{^{-1}}g using Fisher-vector products with H. The resulting direction is scaled to satisfy the KL constraint, and a backtracking line search over step index j decides whether the step is accepted. The policy update uses a maximum KL of 0.01, a Fisher damping of 0.1, and 10 conjugate-gradient iterations. The value function is fit separately to GAE targets (Section VI-B), with 10 epochs of Adam at a learning rate of 1\ensuremath{\times}10\ensuremath{^{-3}} over minibatches of 4,096 samples.

\subsection{Training budget and checkpoint-compatibility layer}

Table VIII lists the training budget and main hyperparameters for each algorithm's final reported run. On the reference machine, these runs took 7.3 min for PPO, 7.4 min for TRPO, 29.0 min for DDPG, 34.4 min for SAC, and 81.8 min for TD3. PPO and TRPO's on-policy rollouts run at about 45,000 environment steps per second on this hardware, roughly 25 to 50 times faster than the three off-policy trainers. This is why PPO and TRPO, despite far larger step budgets (20 million against 1.8 to 9 million), finish in a fraction of the time.

\begin{table*}[!t]
\renewcommand{\thetable}{VIII}
\caption{Training configuration, final accepted run}
\label{tab:VIII}
\centering\footnotesize
\setlength{\tabcolsep}{3pt}
\begin{tabularx}{\textwidth}{@{}>{\raggedright\arraybackslash}X>{\raggedright\arraybackslash}X>{\raggedright\arraybackslash}X>{\raggedright\arraybackslash}X>{\raggedright\arraybackslash}X@{}}
\hline
{\bfseries Hyperparameter} & {\bfseries SAC} & {\bfseries TD3} & {\bfseries DDPG} & {\bfseries TRPO} \\
\hline
Total env-steps & 1,802,240 (early-stopped at 90\%) & 8,994,816 & 2,998,272 & 20,004,864 \\
Discount, \ensuremath{\gamma} & 0.95 & 0.99 & 0.99 & 0.95 \\
Action / exploration noise & entropy-driven + 0.15 rollout noise & \ensuremath{\sigma} = 0.20 & \ensuremath{\sigma} = 0.10 & N/A \\
Target entropy & -1.8 (floor log \ensuremath{\alpha} \ensuremath{\geq} ln 0.018) & N/A & N/A & N/A \\
Action-penalty coefficient & N/A & N/A & N/A & 0.01 \\
Curriculum & Disabled & Disabled & Disabled & Enabled (0.35 fraction) \\
Gradient updates / env. step & 8 & 8 & 8 & N/A (on-policy) \\
Network & MLP, 2\ensuremath{\times}256, tanh & MLP, 2\ensuremath{\times}256, tanh & MLP, 2\ensuremath{\times}256, tanh & MLP, 2\ensuremath{\times}256, tanh \\
\hline
\end{tabularx}
\end{table*}

A shared checkpoint-compatibility layer wraps each off-policy actor network, at save time, into a throwaway instance of PPO's own ActorCritic class, so PPO's evaluate(), evaluate\_multi(), and checkpoint-selection functions load and score any of the four checkpoints without evaluation code changing between algorithms; this is possible because those functions call only a deterministic act(observation) method and never touch the log-standard-deviation or value-function heads the on-policy trainers alone use during training. The same compatibility layer, and the same evaluation functions, score every classical and model-based controller of Section VIII, so every number reported in Section XI is produced by one evaluation codebase regardless of whether the controller behind it was trained or hand-derived.

\section{Classical and Model-Based Control Baselines}

To place the five learned policies of Sections VI and VII alongside the standard tools of marine control engineering, six classical and model-based controllers are implemented on the identical PipelineTrackingEnv guidance signals and thruster allocation of Section V. They are evaluated by exactly the same code as the five reinforcement-learning policies (Section X). None of the six baselines is trained by reinforcement learning. Five are hand-derived or hand-tuned control laws, and the sixth, the adaptive neuro-fuzzy inference system of Section VIII-F, is fit by supervised learning to imitate one of the other five. All six share the guidance geometry, channel definitions, and thruster-allocation map summarized in Table IX and detailed in the closing paragraph of this section. The per-channel gains and structural parameters given for each controller in Sections VIII-A through VIII-F are therefore the only way in which the six baselines differ from one another.

\begin{table}[!t]
\renewcommand{\thetable}{IX}
\caption{Shared guidance and control settings, all six classical and model-based baselines}
\label{tab:IX}
\centering\footnotesize
\setlength{\tabcolsep}{3pt}
\begin{tabularx}{\columnwidth}{@{}>{\raggedright\arraybackslash}X>{\raggedright\arraybackslash}X@{}}
\hline
{\bfseries Parameter} & {\bfseries Value} \\
\hline
Controlled channels & cross\_h (sway), cross\_v (heave/standoff), yaw, roll, pitch, surge \\
Cruise speed (surge setpoint) & 1.825 m/s \\
Standoff target (heave setpoint) & 1.2 m above the pipeline centerline (1.0 m hard floor, Section V-C) \\
Roll / pitch setpoint & 0 rad (active levelling) \\
Heading guidance & Line-of-sight, look-ahead point 1.5 m along the route \\
Maximum episode length & 1,040 steps (Section V-A) \\
Channel output range & Normalized {[}-1, 1{]}, scaled by the per-thruster maximum force T\_max (Section IV-E) \\
Thruster allocation & Moore-Penrose pseudo-inverse of the 6\ensuremath{\times}8 allocation matrix A (Section IV-E) \\
Saturation handling & Uniform scaling of all eight thruster forces, preserving wrench direction (Section IV-E) \\
\hline
\end{tabularx}
\end{table}

\subsection{Proportional-Integral-Derivative (PID) Control}

A standard per-channel proportional-integral-derivative (PID) loop runs directly on the same guidance-law errors that the reinforcement-learning policies observe. These are the horizontal and vertical cross-track/standoff errors, the look-ahead heading error described in the closing paragraph of this section, and the vehicle's roll and pitch angles, which an active leveling term drives toward zero. Each of the six channels (surge, sway, heave, roll, pitch, and yaw) has its own independently tuned proportional, integral, and derivative gains. Integral action is clamped to prevent windup, both during the vehicle's brief startup transient and while a rock obstacle forces a large temporary cross-track error \cite{b27}. The six per-channel PID outputs are combined into a single desired 6-DOF wrench, which the shared allocation described in the closing paragraph below maps to the eight thruster commands. PID is the simplest of the six baselines. Because it uses no model of the vehicle's dynamics, it serves as the reference point against which every other classical and model-based controller in this section is compared.

For a channel with error e and error rate \ensuremath{\dot{e}}, the control law is

\begin{equation}
u = k_{p}e + k_{i}\int e\ dt + k_{d}\dot{e}\tag{44}\label{eq:44}
\end{equation}

where u is the normalized, pre-allocation channel command, k\_p, k\_i, k\_d are the channel's proportional, integral, and derivative gains, and the integral term \ensuremath{\int}e dt is clamped to \ensuremath{\pm}1.0 to limit windup; the resulting u is itself clamped to \ensuremath{\pm}1.0 before being combined into the six-channel wrench of the closing paragraph below.

In continuous time, Equation (44) is the time-domain form of the classical proportional-integral-derivative transfer function

\begin{equation*}
\adjustbox{max width=\columnwidth}{$\displaystyle C(s) = k_{p} + \frac{k_{i}}{s} + k_{d}s$}
\end{equation*}

though the implementation runs the discrete difference form of Equation (44) directly on the guidance error at the controller's 20 Hz update rate, with the integral evaluated by trapezoidal accumulation and the derivative by a backward difference on the logged error signal.

\begin{table}[!t]
\renewcommand{\thetable}{X}
\caption{PID gains, per channel}
\label{tab:X}
\centering\footnotesize
\setlength{\tabcolsep}{3pt}
\begin{tabularx}{\columnwidth}{@{}>{\raggedright\arraybackslash}X>{\raggedright\arraybackslash}X>{\raggedright\arraybackslash}X>{\raggedright\arraybackslash}X@{}}
\hline
{\bfseries Channel} & {\bfseries k\_p} & {\bfseries k\_i} & {\bfseries k\_d} \\
\hline
cross\_h & 0.9 & 0.05 & 0.6 \\
cross\_v & 0.9 & 0.05 & 0.6 \\
yaw & 1.1 & 0.02 & 0.35 \\
roll & 1.4 & 0.0 & 0.30 \\
pitch & 1.4 & 0.0 & 0.30 \\
surge & 0.6 & 0.10 & 0.0 \\
\hline
\end{tabularx}
\par\vspace{3pt}\parbox{\columnwidth}{\scriptsize \emph{Gains for the PID law of Equation (44), tuned independently per channel (Section VIII-A). Roll and pitch carry no integral action, since the active-levelling target is always zero and a small proportional-derivative loop is sufficient to hold it.}}
\end{table}

\subsection{Sliding-Mode Control (SMC)}

Sliding-mode control defines, for each of the six controlled channels, a sliding surface s = \ensuremath{\dot{e}} + \ensuremath{\lambda}e combining the tracking error e and its rate of change \ensuremath{\dot{e}} with a tuned constant \ensuremath{\lambda} setting the desired rate of convergence to the surface \cite{b28,b29}. Rather than a hard sign(s) switching law, which drives the surface to zero in finite time but chatters the thruster commands at the switching frequency, the implemented reaching law replaces sign(s) with a boundary-layer approximation tanh(s/\ensuremath{\varphi}), where \ensuremath{\varphi} is a tuned boundary-layer width; this trades a small, bounded steady-state tracking error for thruster commands that vary smoothly rather than switching at the actuator's command rate, a standard modification for a system, like this one, driven by physical thrusters with finite bandwidth. \ensuremath{\lambda} and \ensuremath{\varphi} are tuned per channel by the same process used for the PID gains of Section VIII-A, and the resulting six-channel wrench is mapped to thruster commands identically to every other baseline in this section.

\begin{equation}
\sigma = \dot{e} + \lambda e\tag{45}\label{eq:45}
\end{equation}

\begin{equation}
u = K\tanh(\sigma/\varphi)\tag{46}\label{eq:46}
\end{equation}

where \ensuremath{\sigma} is the sliding surface, \ensuremath{\lambda} is the surface-slope constant, K is the switching gain, \ensuremath{\varphi} is the boundary-layer width, and u is the resulting normalized channel command, clamped to \ensuremath{\pm}1.0.

The boundary-layer law of Equation (46) approximates the ideal, discontinuous switching law u = K\ensuremath{\cdot}sign(\ensuremath{\sigma}), whose finite-time convergence to the sliding surface follows from the Lyapunov candidate V = \ensuremath{\tfrac{1}{2}}\ensuremath{\sigma}\ensuremath{^{2}} together with the reaching condition

\begin{equation*}
\adjustbox{max width=\columnwidth}{$\displaystyle \dot{V} = \sigma\dot{\sigma} \leq - \eta|\sigma|,\ \ \eta > 0$}
\end{equation*}

which drives \ensuremath{\sigma} to zero in finite time regardless of the specific plant dynamics acting on it \cite{b28}. Replacing sign(\ensuremath{\sigma}) with tanh(\ensuremath{\sigma}/\ensuremath{\varphi}) trades this finite-time guarantee for a small, bounded steady-state tracking error of order \ensuremath{\varphi}, in exchange for a continuous, chatter-free thruster command.

\begin{table}[!t]
\renewcommand{\thetable}{XI}
\caption{Sliding-mode control parameters, per channel}
\label{tab:XI}
\centering\footnotesize
\setlength{\tabcolsep}{3pt}
\begin{tabularx}{\columnwidth}{@{}>{\raggedright\arraybackslash}X>{\raggedright\arraybackslash}X>{\raggedright\arraybackslash}X>{\raggedright\arraybackslash}X@{}}
\hline
{\bfseries Channel} & {\bfseries \ensuremath{\lambda} (surface slope)} & {\bfseries K (switching gain)} & {\bfseries \ensuremath{\varphi} (boundary-layer width)} \\
\hline
cross\_h & 0.5 & 0.35 & 0.6 \\
cross\_v & 0.5 & 0.35 & 0.6 \\
yaw & 0.6 & 0.35 & 0.8 \\
roll & 0.8 & 0.45 & 0.5 \\
pitch & 0.8 & 0.45 & 0.5 \\
surge & 0.4 & 0.35 & 0.6 \\
\hline
\end{tabularx}
\par\vspace{3pt}\parbox{\columnwidth}{\scriptsize \emph{Parameters of the reaching law of Equations (45)-(46) (Section VIII-B). A wider boundary layer, as used for the roll and yaw channels, trades a larger steady-state tracking error for smoother thruster commands.}}
\end{table}

\subsection{Fuzzy Logic Control}

A hand-designed Takagi-Sugeno fuzzy proportional-derivative controller is implemented on each channel. Five triangular membership functions cover the channel's error and five more cover its error-rate, giving a 5\ensuremath{\times}5, 25-rule table per channel \cite{b30}. Each rule's firing strength is the product t-norm of its error and error-rate membership degrees. The controller output is the Sugeno-style weighted average of the rules' linear consequents, each weighted by its firing strength, so no defuzzification over a fuzzy output set is needed. The rule base was designed once, by inspecting the same error and error-rate ranges used to tune the PID gains of Section VIII-A, and it is not adapted online. Section VIII-F returns to a similar error/error-rate mapping in a learned form.

The channel's error and error-rate are first scaled and clipped to the controller's normalized universe of discourse,

\begin{equation}
\widehat{e} = \operatorname{clip}\left( G_{e}e, - 1,1 \right),\ \ \widehat{\dot{e}} = \operatorname{clip}\left( G_{\dot{e}}\dot{e}, - 1,1 \right)\tag{47}\label{eq:47}
\end{equation}

each covered by five triangular membership functions centred at -1, -0.5, 0, 0.5, and 1 with half-width 0.5, labelled negative-big (NB), negative-small (NS), zero (Z), positive-small (PS), and positive-big (PB). Each triangular set, with centre c and half-width b, has the piecewise-linear membership degree

\begin{equation*}
\adjustbox{max width=\columnwidth}{$\displaystyle \mu(x) = \operatorname{max}\left( 0,\ 1 - \frac{|x - c|}{b} \right)$}
\end{equation*}

equal to 1 at the centre and falling linearly to 0 at a distance b on either side, giving the five overlapping sets above their explicit analytic form. Writing \ensuremath{\mu}\_i for the membership degree of \ensuremath{\hat{e}} in its i-th triangular set and \ensuremath{\mu}\_j for the membership degree of \ensuremath{\hat{e}\ensuremath{\dot{}}} in its j-th set, rule (i,j)'s firing strength uses the product t-norm

\begin{equation}
w_{ij} = \mu_{i}\left( \widehat{e} \right)\mu_{j}\left( \widehat{\dot{e}} \right)\tag{48}\label{eq:48}
\end{equation}

and the controller's output is the Sugeno-style weighted average of the 25 rules' constant consequents y\_ij, listed in Table XIII, scaled by an output gain G\_u:

\begin{equation}
u = G_{u}\frac{\sum_{i,j}^{}{w_{ij}y_{ij}}}{\sum_{i,j}^{}w_{ij}}\tag{49}\label{eq:49}
\end{equation}

where G\_e, G\_\ensuremath{\dot{e}}, and G\_u are the channel's error, error-rate, and output scaling gains, \ensuremath{\hat{e}} and \ensuremath{\hat{e}\ensuremath{\dot{}}} are the scaled and clipped inputs, w\_ij is rule (i,j)'s firing strength, y\_ij is that rule's constant consequent, and u is the resulting normalized channel command, clamped to \ensuremath{\pm}1.0.

\begin{table}[!t]
\renewcommand{\thetable}{XII}
\caption{Fuzzy-logic scaling gains, per channel}
\label{tab:XII}
\centering\footnotesize
\setlength{\tabcolsep}{3pt}
\begin{tabularx}{\columnwidth}{@{}>{\raggedright\arraybackslash}X>{\raggedright\arraybackslash}X>{\raggedright\arraybackslash}X>{\raggedright\arraybackslash}X@{}}
\hline
{\bfseries Channel} & {\bfseries G\_e (error scale)} & {\bfseries G\_\ensuremath{\dot{e}} (error-rate scale)} & {\bfseries G\_u (output scale)} \\
\hline
cross\_h & 1.2 & 0.8 & 1.0 \\
cross\_v & 1.2 & 0.8 & 1.0 \\
yaw & 1.5 & 0.6 & 1.2 \\
roll & 2.0 & 0.8 & 1.5 \\
pitch & 2.0 & 0.8 & 1.5 \\
surge & 1.0 & 0.0 & 0.8 \\
\hline
\end{tabularx}
\par\vspace{3pt}\parbox{\columnwidth}{\scriptsize \emph{Gains entering Equation (47) and the output scaling of Equation (49) (Section VIII-C). The surge channel carries no error-rate scaling, since only the error term is used to regulate cruise speed.}}
\end{table}

\begin{table*}[!t]
\renewcommand{\thetable}{XIII}
\caption{Fuzzy-logic rule table (shared across all six channels): consequent y\_ij for error i and error-rate j}
\label{tab:XIII}
\centering\footnotesize
\setlength{\tabcolsep}{3pt}
\begin{tabularx}{\textwidth}{@{}>{\raggedright\arraybackslash}X>{\raggedright\arraybackslash}X>{\raggedright\arraybackslash}X>{\raggedright\arraybackslash}X>{\raggedright\arraybackslash}X>{\raggedright\arraybackslash}X@{}}
\hline
{\bfseries e \textbackslash{} \ensuremath{\dot{e}}} & {\bfseries NB} & {\bfseries NS} & {\bfseries Z} & {\bfseries PS} & {\bfseries PB} \\
\hline
NB & -1.0 & -1.0 & -1.0 & -0.5 & 0.0 \\
NS & -1.0 & -1.0 & -0.5 & 0.0 & 0.5 \\
Z & -1.0 & -0.5 & 0.0 & 0.5 & 1.0 \\
PS & -0.5 & 0.0 & 0.5 & 1.0 & 1.0 \\
PB & 0.0 & 0.5 & 1.0 & 1.0 & 1.0 \\
\hline
\end{tabularx}
\par\vspace{3pt}\parbox{\textwidth}{\scriptsize \emph{The same 25-value rule table of Equation (49), before the per-channel scaling of Table XII, is reused unmodified across all six channels; the anti-symmetric structure runs from full negative output when both error and error-rate are strongly negative to full positive output when both are strongly positive, with a zero-output ridge along the anti-diagonal where the two disagree.}}
\end{table*}

\subsection{Feedback Linearization (FL)}

Feedback linearization is a model-based computed-torque controller \cite{b28}. It has privileged access to the same Fossen dynamics model that the simulator uses to propagate the vehicle, including the mass matrix, Coriolis-centripetal coupling (Section IV-B), damping, and restoring forces. At every control step, the controller evaluates these nonlinear terms at the vehicle's current state and commands a wrench that cancels them exactly. The closed loop then reduces to a linear, decoupled error-dynamics system, driven by a desired acceleration that simple proportional-derivative gains set from the guidance-law error. Because the nonlinear dynamics are cancelled algebraically and not approximated, the outer PD gains can be tuned independently on each channel, free of the cross-coupling that complicates gain tuning for classical PID and SMC. By construction, feedback linearization is the cleanest and fastest of the six baselines to tune, since it needs no boundary-layer, membership-function, or rule-base design. Its cost is a dependence on an accurate dynamics model, which purely error-driven controllers such as PID and SMC do not need.

A per-channel desired acceleration is formed from the guidance error exactly as for PID, but without an integral term, since the model cancellation below removes the steady-state bias an integrator would otherwise be needed to correct:

\begin{equation}
a = k_{p}e + k_{d}\dot{e}\tag{50}\label{eq:50}
\end{equation}

and the commanded wrench cancels the governing equation's nonlinear terms (Equation 1) about the vehicle's current state while imposing this desired acceleration on the mass matrix:

\begin{equation}
\tau = Ma + C(\nu)\nu + D(\nu) + g(\eta)\tag{51}\label{eq:51}
\end{equation}

where a is the desired body-frame acceleration, k\_p and k\_d are the channel's proportional and derivative gains, \ensuremath{\tau} is the resulting commanded wrench, and M, C(\ensuremath{\nu})\ensuremath{\nu}, D(\ensuremath{\nu}), and g(\ensuremath{\eta}) are evaluated at the vehicle's current state exactly as in the governing equation of Section IV-A; substituting Equation (50) into Equation (1) gives M\ensuremath{\dot{\ensuremath{\nu}}} = Ma, so the closed-loop channel dynamics reduce exactly to \ensuremath{\dot{\ensuremath{\nu}}} = a, a linear, decoupled system driven by the proportional-derivative law of Equation (50). Writing this out for a constant setpoint, so that the reference acceleration is zero, the tracking error of every channel then obeys the linear, second-order ordinary differential equation

\begin{equation*}
\adjustbox{max width=\columnwidth}{$\displaystyle \ddot{e} + k_{d}\dot{e} + k_{p}e = 0$}
\end{equation*}

which is critically or over-damped whenever k\_d\ensuremath{^{2}} \ensuremath{\geq} 4k\_p; the gains of Table XIV can therefore be chosen directly from a target natural frequency \ensuremath{\omega}\_n = \ensuremath{\surd}k\_p and damping ratio \ensuremath{\zeta} = k\_d/(2\ensuremath{\surd}k\_p) rather than by trial and error, a design freedom exact cancellation of the nonlinear terms makes possible.

\begin{table}[!t]
\renewcommand{\thetable}{XIV}
\caption{Feedback-linearization gains, per channel}
\label{tab:XIV}
\centering\footnotesize
\setlength{\tabcolsep}{3pt}
\begin{tabularx}{\columnwidth}{@{}>{\raggedright\arraybackslash}X>{\raggedright\arraybackslash}X>{\raggedright\arraybackslash}X@{}}
\hline
{\bfseries Channel} & {\bfseries k\_p (1/s\ensuremath{^{2}})} & {\bfseries k\_d (1/s)} \\
\hline
cross\_h & 3.0 & 2.5 \\
cross\_v & 3.0 & 2.5 \\
yaw & 4.0 & 2.0 \\
roll & 6.0 & 2.5 \\
pitch & 6.0 & 2.5 \\
surge & 2.0 & 0.0 \\
\hline
\end{tabularx}
\par\vspace{3pt}\parbox{\columnwidth}{\scriptsize \emph{Gains of the desired-acceleration law, Equation (50) (Section VIII-D), in acceleration units rather than the normalized units of Tables X-XIII, since the nonlinear cancellation of Equation (51) restores physical units before the result is mapped to thruster commands.}}
\end{table}

\subsection{Model Predictive Control (MPC)}

The model predictive controller (MPC) replans a 20-step action sequence at every control step using the cross-entropy method (CEM) \cite{b31}. A population of candidate sequences is rolled out through the same Fossen model used for feedback linearization, and each candidate is scored by a quadratic tracking-and-effort cost over the horizon. Over two CEM iterations, a Gaussian action distribution is refit to the best-scoring elite fraction of the previous iteration's samples before the population is resampled. A common practice is to seed the first CEM iteration from the previous step's plan. Here, each replanning step is instead warm-started from a fresh feedback-linearization trajectory (Section VIII-D) computed over the same horizon, which gives the sampler a strong, dynamically consistent starting point in place of an open-loop guess. In standard receding-horizon fashion, only the first action of the resulting plan is applied before the full horizon is replanned. The executed action, however, departs from the usual choice of the elite-mean sequence. It is the first action of the single highest-scoring sampled sequence, so the controller commits to a candidate that was actually evaluated and never to an averaged sequence that no individual sample achieved.

In its general form, the receding-horizon controller solves, at every control step, the finite-horizon optimal control problem

\begin{equation*}
\adjustbox{max width=\columnwidth}{$\displaystyle U^{*} = arg\min_{U = \{ u_{0},\ldots,u_{H - 1}\}}J(U)\ \ \text{s.t.}\ \ x_{k + 1} = f\left( x_{k},u_{k} \right),\ u_{k} \in \lbrack - 1,1\rbrack^{8},\ x_{0} = x(t)$}
\end{equation*}

of which the cross-entropy method used here is one derivative-free way of approximating the minimizing sequence U* by sampling rather than by gradient-based optimization, appropriate given that the flip and pipe-gap-violation events the cost of Equation (52) does not otherwise penalize smoothly are discontinuous in the action sequence.

Each candidate H-step sequence U = \{u\_0,\ldots,u\_\{H-1\}\} is scored by

\begin{equation}
\begin{split}J(U) = \sum_{k=0}^{H-1}\Big[&W_{CROSS}\,cross_{h,k}^{2} + W_{STANDOFF}\,e_{v,k}^{2}\\
&+ W_{YAW}\,e_{\psi,k}^{2} + W_{ATTITUDE}\left(\varphi_{k}^{2} + \theta_{k}^{2}\right)\\
&+ W_{SURGE}\,e_{u,k}^{2} + W_{EFFORT}\sum u_{k}^{2}\Big]\end{split}\tag{52}\label{eq:52}
\end{equation}

and the CEM population mean and standard deviation are refit from the E best-scoring, or elite, sampled sequences before the next of the two resampling iterations:

\begin{equation}
\mu \leftarrow \operatorname{mean}\left( U_{elite} \right),\ \ \sigma \leftarrow std\left( U_{elite} \right)\tag{53}\label{eq:53}
\end{equation}

with the action ultimately executed being the first action of the single best-scoring sequence found across both iterations rather than of the elite mean,

\begin{equation}
u_{t}^{*} = U_{best}\lbrack 0\rbrack,\ \ U_{best} = arg\min_{U \in samples}J(U)\tag{54}\label{eq:54}
\end{equation}

and the surge setpoint entering the W\_SURGE term of Equation (52) adapts to the vehicle's current heading error rather than remaining fixed at the cruise speed of Table IX:

\begin{equation}
u_{surge,sp} = 1.825 \cdot \operatorname{clip}\left( \cos\left( e_{\psi} \right),0.2,1.0 \right)\tag{55}\label{eq:55}
\end{equation}

where cross\_h,k, e\_v,k, e\_\ensuremath{\psi},k, \ensuremath{\varphi}\_k, \ensuremath{\theta}\_k, and e\_u,k are, respectively, the predicted horizontal cross-track, vertical standoff, yaw, roll, pitch, and surge errors at step k of the candidate sequence; W\_CROSS, W\_STANDOFF, W\_YAW, W\_ATTITUDE, W\_SURGE, and W\_EFFORT are the corresponding cost weights of Table XV; \ensuremath{\mu} and \ensuremath{\sigma} are the CEM sampling distribution's mean and standard deviation; U\_elite is the elite fraction of scored candidate sequences; U\_best is the single best-scoring sequence found; u*\_t is the action applied at the current step; and u\_surge,sp is the adaptive surge setpoint, which reduces the commanded cruise speed as the vehicle's heading error away from the look-ahead point grows, so the controller slows rather than overshoots while turning through the route's corner.

\begin{table}[!t]
\renewcommand{\thetable}{XV}
\caption{Model predictive control parameters}
\label{tab:XV}
\centering\footnotesize
\setlength{\tabcolsep}{3pt}
\begin{tabularx}{\columnwidth}{@{}>{\raggedright\arraybackslash}X>{\raggedright\arraybackslash}X@{}}
\hline
{\bfseries Parameter} & {\bfseries Value} \\
\hline
Prediction horizon, H & 20 steps \\
Samples per CEM iteration, K & 256 \\
CEM iterations & 2 \\
Elite fraction & 0.1 (minimum 4 elites; 25 in practice) \\
Sampling noise, initial standard deviation & 0.5 \\
Action bounds & {[}-1, 1{]} per thruster (8 thrusters) \\
W\_CROSS (horizontal cross-track\ensuremath{^{2}}) & 6.0 \\
W\_STANDOFF (vertical standoff error\ensuremath{^{2}}) & 5.0 \\
W\_YAW (yaw error\ensuremath{^{2}}) & 3.0 \\
W\_ATTITUDE (roll\ensuremath{^{2}} + pitch\ensuremath{^{2}}) & 2.0 \\
W\_SURGE (surge error\ensuremath{^{2}}) & 3.0 \\
W\_EFFORT (\ensuremath{\Sigma} thrust\ensuremath{^{2}}) & 0.02 \\
Heading look-ahead & 1.5 m (a 3.0 m look-ahead was tried and caused corner overshoot) \\
Sampling-distribution seed & Feedback-linearization law of Section VIII-D, re-computed every control step \\
Executed action & First action of the single best-ever-scoring sampled sequence (Equation 54), not the elite mean \\
Prediction model & The same Fossen dynamics model used by the simulator (Section IV) \\
\hline
\end{tabularx}
\par\vspace{3pt}\parbox{\columnwidth}{\scriptsize \emph{Parameters of the cost function of Equation (52) and the cross-entropy-method search of Equations (53)-(55) (Section VIII-E).}}
\end{table}

\subsection{Adaptive Neuro-Fuzzy Inference System (ANFIS)}

The sixth baseline is a Sugeno-type adaptive neuro-fuzzy inference system, implemented per channel as a small network of Gaussian membership functions with learned centre and width parameters, feeding a layer of rule-firing strengths and a learned linear consequent per rule \cite{b32}, structurally the same rule-table shape as the hand-designed fuzzy controller of Section VIII-C but with every membership-function and consequent parameter free to be learned rather than hand-set. Rather than independently tuning this network against the tracking task directly, it is trained by supervised gradient descent to behaviour-clone the PID controller of Section VIII-A: the network is fit, offline, to reproduce the PID law's own (error, error-rate) \ensuremath{\rightarrow} output mapping as closely as possible, so that ANFIS in this study is best read as a compact, learned surrogate that approximates the PID controller rather than as an independently-optimized sixth control law, and its results in Section XI should be interpreted as an upper bound on how closely a small learned network can reproduce PID's behaviour rather than as evidence of an independently superior fuzzy-neural controller.

Each channel takes two raw inputs, the error e and error-rate \ensuremath{\dot{e}}, each covered by three learned Gaussian membership functions,

\begin{equation}
\mu_{m}(x) = \operatorname{exp}\left( - \frac{\left( x - c_{m} \right)^{2}}{2{\sigma_{m}}^{2}} \right),\ \ m = 1,2,3\tag{56}\label{eq:56}
\end{equation}

giving nine rules, one per combination of an error membership function and an error-rate membership function, whose firing strength again uses the product t-norm,

\begin{equation}
w_{r} = \mu_{i}(e)\mu_{j}\left( \dot{e} \right),\ \ r = 3i + j\tag{57}\label{eq:57}
\end{equation}

normalized across the nine rules,

\begin{equation}
{\overline{w}}_{r} = w_{r}/\sum_{r = 1}^{9}w_{r}\tag{58}\label{eq:58}
\end{equation}

and combined with each rule's own learned, constant (zero-order Sugeno) consequent p\_r into the channel output. Cast in the standard five-layer ANFIS architecture \cite{b32}, Layer 1 performs the Gaussian fuzzification of Equation (56), Layer 2 the rule-firing strength of Equation (57), and Layer 3 the normalization of Equation (58); Layer 4 then forms each rule's individual weighted output,

\begin{equation*}
\adjustbox{max width=\columnwidth}{$\displaystyle f_{r} = {\overline{w}}_{r}p_{r}$}
\end{equation*}

and Layer 5 sums the nine layer-4 outputs into the network output,

\begin{equation}
u = \sum_{r = 1}^{9}{{\overline{w}}_{r}p_{r}}\tag{59}\label{eq:59}
\end{equation}

where c\_m and \ensuremath{\sigma}\_m are the m-th membership function's learned centre and width for a given input, w\_r is rule r's firing strength, \ensuremath{\bar{w}}\_r is its normalized firing strength, p\_r is rule r's learned constant consequent, and u is the resulting normalized channel command, clamped to \ensuremath{\pm}1.0.

All membership-function and consequent parameters, Tables XVI and XVII, are learned jointly by minimizing the mean-squared error between the network's output and the PID controller's output over a fixed dataset of 15 behaviour-cloning episodes,

\begin{equation}
L = \operatorname{mean}\left( \left( u_{ANFIS} - u_{PID} \right)^{2} \right)\tag{60}\label{eq:60}
\end{equation}

optimized by Adam at a learning rate of 0.03 for 400 epochs.

\begin{table*}[!t]
\renewcommand{\thetable}{XVI}
\caption{ANFIS trained membership-function parameters, per channel (centre c, width \ensuremath{\sigma}; three membership functions each for e and \ensuremath{\dot{e}})}
\label{tab:XVI}
\centering\footnotesize
\setlength{\tabcolsep}{3pt}
\begin{tabularx}{\textwidth}{@{}>{\raggedright\arraybackslash}X>{\raggedright\arraybackslash}X>{\raggedright\arraybackslash}X>{\raggedright\arraybackslash}X>{\raggedright\arraybackslash}X@{}}
\hline
{\bfseries Channel} & {\bfseries c\_e (MF 1, 2, 3)} & {\bfseries \ensuremath{\sigma}\_e (MF 1, 2, 3)} & {\bfseries c\_\ensuremath{\dot{e}} (MF 1, 2, 3)} & {\bfseries \ensuremath{\sigma}\_\ensuremath{\dot{e}} (MF 1, 2, 3)} \\
\hline
cross\_h & -1.189, -0.837, 0.763 & 0.535, 0.605, 0.693 & -1.130, 0.389, 1.155 & 0.753, 0.705, 0.535 \\
cross\_v & -1.012, -0.585, 1.083 & 0.733, 0.668, 0.777 & -1.165, 1.263, 1.048 & 0.634, 0.312, 0.693 \\
yaw & -1.055, -0.520, 0.751 & 1.021, 0.048, 1.010 & -2.145, -0.610, 1.623 & 0.292, 1.500, 1.528 \\
roll & -1.242, -0.595, 1.098 & 0.665, 0.368, 0.748 & -1.013, 0.828, 1.023 & 0.886, 0.376, 0.839 \\
pitch & -1.316, -0.726, 1.230 & 0.534, 0.318, 0.567 & -1.107, 0.059, 1.205 & 0.824, 0.588, 0.703 \\
surge & -1.722, -0.252, 2.165 & 0.484, 0.735, 0.580 & -0.435, 0.000, 0.435 & 1.494, 1.001, 1.494 \\
\hline
\end{tabularx}
\par\vspace{3pt}\parbox{\textwidth}{\scriptsize \emph{Learned parameters of Equation (56), trained by the behaviour-cloning loss of Equation (60) (Section VIII-F). Unlike the hand-designed fuzzy controller's symmetric, evenly-spaced membership functions (Section VIII-C), these centres and widths are free to settle asymmetrically wherever the training data best supports them.}}
\end{table*}

\begin{table*}[!t]
\renewcommand{\thetable}{XVII}
\caption{ANFIS trained rule consequents, per channel (rule r = 3i + j, i indexing the error membership function, j the error-rate membership function)}
\label{tab:XVII}
\centering\footnotesize
\setlength{\tabcolsep}{3pt}
\begin{tabularx}{\textwidth}{@{}>{\raggedright\arraybackslash}X>{\raggedright\arraybackslash}X>{\raggedright\arraybackslash}X>{\raggedright\arraybackslash}X>{\raggedright\arraybackslash}X>{\raggedright\arraybackslash}X>{\raggedright\arraybackslash}X>{\raggedright\arraybackslash}X>{\raggedright\arraybackslash}X>{\raggedright\arraybackslash}X@{}}
\hline
{\bfseries Channel} & {\bfseries r1} & {\bfseries r2} & {\bfseries r3} & {\bfseries r4} & {\bfseries r5} & {\bfseries r6} & {\bfseries r7} & {\bfseries r8} & {\bfseries r9} \\
\hline
cross\_h & -1.009 & -0.798 & -0.373 & -1.067 & -0.233 & 0.280 & -0.345 & 0.583 & 0.626 \\
cross\_v & -0.665 & -0.023 & -0.215 & -0.400 & 0.097 & -0.081 & 0.495 & 0.843 & 1.170 \\
yaw & -1.816 & -2.479 & -0.467 & -0.900 & -0.821 & 0.297 & -0.159 & 0.710 & 2.162 \\
roll & -0.262 & 0.115 & 0.118 & -0.329 & 0.076 & 0.022 & -0.049 & 0.227 & 0.351 \\
pitch & -0.537 & -0.215 & 0.160 & -0.496 & -0.235 & 0.174 & -0.125 & 0.398 & 0.729 \\
surge & 0.545 & 0.579 & 0.545 & 0.520 & 0.505 & 0.520 & 1.091 & 0.877 & 1.091 \\
\hline
\end{tabularx}
\par\vspace{3pt}\parbox{\textwidth}{\scriptsize \emph{Learned constant consequents p\_r of Equation (59) (Section VIII-F). The yaw channel's consequents span the widest range, from -2.479 to 2.162, consistent with yaw carrying the largest error-rate membership-function width (\ensuremath{\sigma}\_\ensuremath{\dot{e}} up to 1.528 in Table XVI) among the six channels.}}
\end{table*}

All six controllers of this section observe the identical guidance geometry used by the five reinforcement-learning policies of Sections VI and VII: a signed horizontal and vertical cross-track/standoff error computed against the nearest point of the piecewise-linear pipeline centerline, a heading error measured against a look-ahead point 1.5 m further along the route rather than against the nearest point's own local tangent, since the tangent direction is discontinuous at the route's corner vertex and a look-ahead point avoids commanding a heading discontinuity there, and an active roll and pitch leveling term driving both angles toward zero. Every controller's desired 6-DOF wrench is mapped to the eight thruster commands by the Moore-Penrose pseudo-inverse of the identical allocation matrix A of Section IV-E, f = A\ensuremath{^{+}}\ensuremath{\tau}\_desired, with any resulting thruster command that exceeds the \ensuremath{\pm}1 normalized range scaled back uniformly across all eight thrusters rather than clipped per-thruster, preserving the commanded wrench's direction at the cost of its magnitude. Because guidance geometry, thruster allocation, and saturation handling are identical for every controller implemented in this study, any performance difference reported in Section XI reflects the control law itself rather than a difference in how error is measured or how a desired wrench becomes a thruster command.

\section{Deployment Inside NVIDIA Isaac Sim}

Once training has produced a checkpoint, or a classical controller's gains have been tuned, the same USD scene is opened inside Isaac Sim and a controller script is attached to the vehicle prim. At every physics step, the script reads the current world pose and rotation matrix from the rigid body and rotates the world-frame linear and angular velocities into the body frame to obtain \ensuremath{\nu}. From \ensuremath{\nu}, it computes the added-mass Coriolis-centripetal correction (Section IX-B) and the damping (Section IV-C). The thruster wrench follows from the current thrust command and the allocation matrix (Section IV-E). Because PhysX gravity is disabled for this body, the restoring force (Section IV-D) is also computed in the script and rotated into the body frame. The Coriolis correction, damping, thrust, and buoyancy terms are then summed into a single body-frame force and torque. PhysX integrates the resulting motion, so no second Runge-Kutta solver runs alongside it.

\subsection{Timestep and added-mass reconciliation}

Added mass is applied by folding it into an effective mass matrix, M\_full\_inv = (M\_RB + M\_A)\ensuremath{^{-1}}, so the externally applied force representing the rigid-body dynamics, F\_ext = M\_RB \ensuremath{\cdot} \ensuremath{\dot{\ensuremath{\nu}}}, causes PhysX's own integrator to reproduce the full, added-mass-corrected acceleration algebraically; the standalone Runge-Kutta model remains the physical source of truth for training and for the tracking-error numbers in Section XI. The policy or controller was tuned at 0.05 s (20 Hz) while Isaac Sim's physics scene runs at its default 60 Hz, so the deployment script queries it once every three physics steps and holds the resulting command constant in between (zero-order hold). Minimum standoff distance and maximum absolute roll or pitch are logged across each run, and the run stops on a detected flip or on reaching the end of the pipeline.

\subsection{The Coriolis-centripetal term in the live controller}

The Coriolis-centripetal term C(\ensuremath{\nu})\ensuremath{\nu} of Section IV-B is included in both the training-side Fossen model and the live Isaac Sim controller, but is not applied identically in the two branches, because the two branches integrate motion by different means. The training-side Runge-Kutta integrator has no notion of rigid-body dynamics beyond the explicit right-hand side supplied to it, so the full term, C\_RB(\ensuremath{\nu})\ensuremath{\nu} plus C\_A(\ensuremath{\nu})\ensuremath{\nu}, is computed and applied there. In the live Isaac Sim controller, only the added-mass Coriolis correction C\_A(\ensuremath{\nu})\ensuremath{\nu} is applied explicitly, because PhysX's own rigid-body solver already supplies the rigid-body Coriolis term C\_RB(\ensuremath{\nu})\ensuremath{\nu} intrinsically as part of its native integration of a rotating rigid body; applying it a second time, on top of PhysX's own treatment of rigid-body rotation, would double-count it and inject a spurious torque with no physical counterpart.

The rigid-body part of the Coriolis effect must be counted exactly once. The training-side model applies it directly, while in live deployment PhysX already accounts for it as it moves the vehicle, so the Isaac Sim controller adds only the added-mass part. Both setups therefore run on the same equations of motion. The ablation in Section XI-A, carried out on the training-side model with both Coriolis components applied explicitly, is what shows that including the full Coriolis term is safe for PPO and TRPO. The deployment split described here simply ensures that the model tested in that ablation matches the one that actually drives the vehicle in Isaac Sim. It is a detail of how the motion is integrated and makes no separate claim of its own.

\section{Evaluation Protocol}

Two evaluation phases are run from each algorithm's trained checkpoint or each classical controller's tuned parameters, both using the standalone Fossen model rather than the interactive Isaac Sim scene, and every controller compared in this paper is evaluated by exactly the same code via the checkpoint-compatibility layer of Section VII-F.

\subsection{Canonical episode and multi-episode statistics}

A single deterministic episode is run with the rocks fixed at their scene positions, logging position, orientation, velocity, action, and reward at every step. The same deterministic controller is then run for 30 further episodes, each with its own domain-randomized rock layout and spawn jitter (Section V-D). A 95 percent Wilson confidence interval is computed for the success, flip, and pipe-gap-violation rates:

\begin{equation}
{\widehat{p}}_{s} = successes/n\tag{61}\label{eq:61}
\end{equation}

\begin{equation}
denom = 1 + \frac{z^{2}}{n},\ centre = {\widehat{p}}_{s} + \frac{z^{2}}{2n}\tag{62}\label{eq:62}
\end{equation}

\begin{equation}
{half}_{w} = z\sqrt{\frac{{\widehat{p}}_{s}\left( 1 - {\widehat{p}}_{s} \right)}{n} + \frac{z^{2}}{4n^{2}}}\tag{63}\label{eq:63}
\end{equation}

\begin{equation}
CI = \left( \frac{centre - {half}_{w}}{denom},\ \frac{centre + {half}_{w}}{denom} \right),\ z = 1.96\tag{64}\label{eq:64}
\end{equation}

where \ensuremath{\hat{p}}\_s is the observed rate, success, flip, or gap-violation, over n episodes, z = 1.96 is the standard-normal critical value for a 95 percent interval, and CI is the resulting Wilson score interval.

\subsection{Tracking error and additional metrics}

A separate script re-derives tracking-error statistics directly from the canonical episode's logged trajectory, independent of the training and evaluation code, using the three-dimensional Euclidean distance from the vehicle's position to the nearest point on the piecewise-linear pipeline centerline:

\begin{equation}
\adjustbox{max width=\columnwidth}{$\displaystyle RMSE = \sqrt{\operatorname{mean}\left( {e_{i}}^{2} \right)},\ MAE = \operatorname{mean}\left( \left| e_{i} \right| \right),\ e_{\max} = \operatorname{max}\left( e_{i} \right)$}\tag{65}\label{eq:65}
\end{equation}

where e\_i is the three-dimensional Euclidean distance from the vehicle's logged position to the nearest point on the pipeline centerline at sample i, and RMSE, MAE, and e\_max are the root-mean-square, mean-absolute, and maximum tracking error over the episode.

If this script's output agrees with the evaluation summary from the training script, that confirms two separately written pieces of code are measuring the same thing. The same logs also give two further metrics for every controller. The first is an energy proxy: the average of the squared normalized thruster command over an episode, which relates to the mean action-magnitude column in Section XI's main comparison table. The second applies only to the five reinforcement-learning algorithms. It measures convergence as the number of environment steps, and the wall-clock time, needed before the deterministic evaluation policy first reaches 90 percent success during training.

\section{Results}

\subsection{Coriolis-term ablation (PPO and TRPO)}

Before the full set of controllers is compared, the Coriolis-centripetal term introduced in Section IV-B is validated through a controlled ablation restricted to PPO and TRPO (Fig. 5). Each controller is evaluated over 30 episodes. In every episode, the ROV starts at the beginning of the pipeline and attempts to follow it to the end. An episode counts as a success only if the vehicle reaches within 0.30 m of the route's final point before any failure occurs. It counts as a failure if the vehicle collides with a rock or the seafloor, comes closer than 1.0 m to the pipe, flips, leaves the permitted operating area, or runs out of time. Each algorithm was trained twice, once with C(\ensuremath{\nu})\ensuremath{\nu} included in the Fossen model and once with it removed, while every other environment, reward, and training setting was held fixed. The four resulting checkpoints were then scored with the identical 30-episode randomized evaluation protocol of Section X. TD3, DDPG, and SAC were also examined informally while this ablation was developed, but their results are not reported numerically here. The comparison is intended as a focused before-and-after test on the two algorithms for which the released Isaac Sim controller of Section IX-B applies the term, and results from three further algorithms would not sharpen that specific comparison.

\begin{table*}[!t]
\renewcommand{\thetable}{XVIII}
\caption{Coriolis-term ablation, PPO and TRPO (30-episode randomized evaluation)}
\label{tab:XVIII}
\centering\footnotesize
\setlength{\tabcolsep}{3pt}
\begin{tabularx}{\textwidth}{@{}>{\raggedright\arraybackslash}X>{\raggedright\arraybackslash}X>{\raggedright\arraybackslash}X>{\raggedright\arraybackslash}X>{\raggedright\arraybackslash}X>{\raggedright\arraybackslash}X@{}}
\hline
{\bfseries Algorithm} & {\bfseries Coriolis term} & {\bfseries Success rate} & {\bfseries Flip rate} & {\bfseries Corridor-gap violation} & {\bfseries Standoff RMSE (m)} \\
\hline
PPO & Without & 100\% & 0\% & 0\% & 0.152 \\
PPO & With & 100\% & 0\% & 0\% & 0.146 \\
TRPO & Without & 100\% & 0\% & 0\% & 0.176 \\
TRPO & With & 100\% & 0\% & 0\% & 0.142 \\
\hline
\end{tabularx}
\end{table*}

\begin{figure*}[!t]
\renewcommand{\thefigure}{5}
\centering
\includegraphics[width=0.9\textwidth,height=0.8\textheight,keepaspectratio]{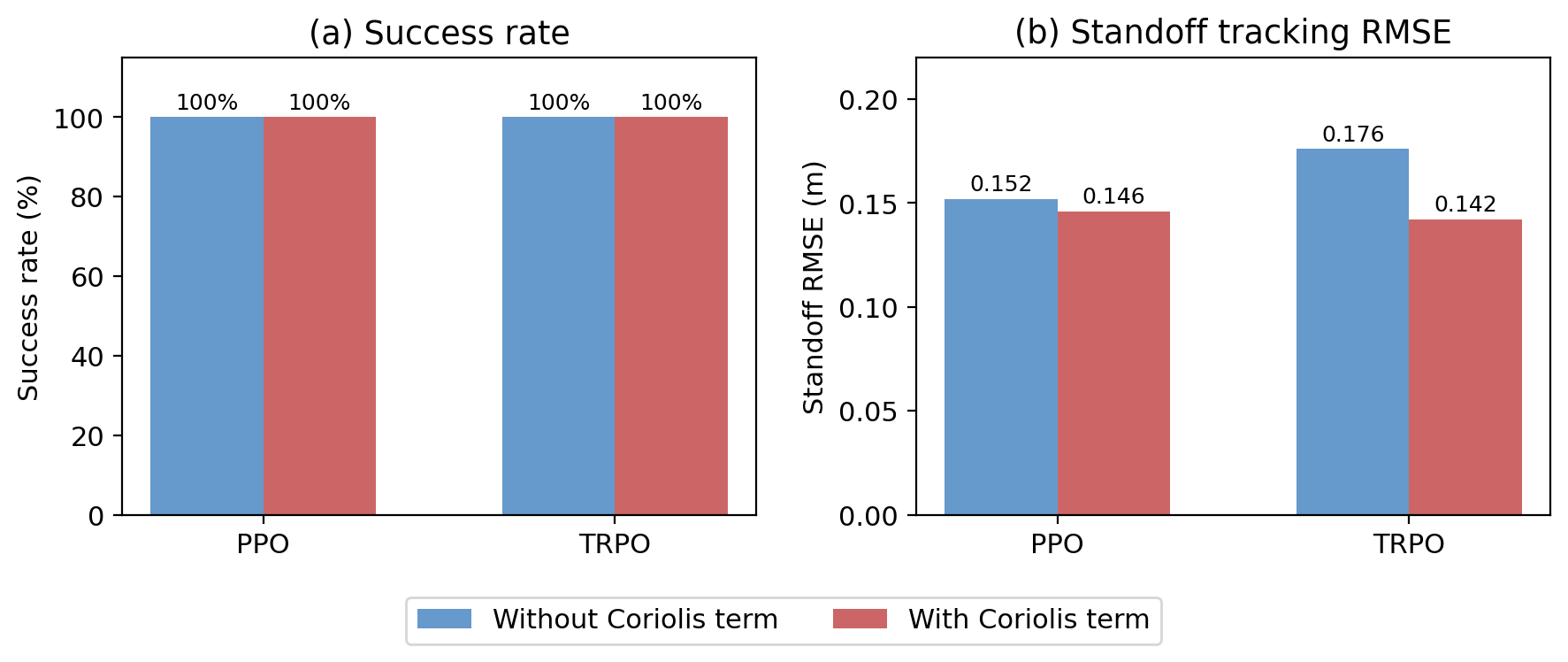}
\caption{Coriolis-term ablation for PPO and TRPO: (a) success rate and (b) standoff root-mean-square error, with and without the Coriolis-centripetal term.}
\label{fig:5}
\end{figure*}

With the Coriolis term included, both algorithms keep a 100 percent success rate with zero flips and zero corridor-gap violations, and both track more accurately. PPO's standoff RMSE falls from 0.152 m to 0.146 m. TRPO's falls from 0.176 m to 0.142 m, a reduction of approximately 19 percent. Adding Coriolis-centripetal coupling therefore leaves both algorithms stable and improves their tracking. On this basis, the full coupling is treated as the primary dynamics model for the rest of this paper, with no simplification needed to work around it.

\subsection{Reinforcement learning versus classical and model-based control}

Table XIX compares ten controllers under the Coriolis-enabled dynamics validated in Section XI-A: four reinforcement learning policies of Sections VI and VII (PPO, TD3, DDPG, and TRPO) and the six classical and model-based baselines of Section VIII (PID, SMC, Fuzzy, FL, MPC, and ANFIS), scored by the identical 30-episode randomized evaluation protocol of Section X.

\begin{table*}[!t]
\renewcommand{\thetable}{XIX}
\caption{Reinforcement learning versus classical and model-based control, Coriolis-enabled dynamics (30-episode randomized evaluation)}
\label{tab:XIX}
\centering\footnotesize
\setlength{\tabcolsep}{3pt}
\begin{tabularx}{\textwidth}{@{}>{\raggedright\arraybackslash}X>{\raggedright\arraybackslash}X>{\raggedright\arraybackslash}X>{\raggedright\arraybackslash}X>{\raggedright\arraybackslash}X>{\raggedright\arraybackslash}X>{\raggedright\arraybackslash}X>{\raggedright\arraybackslash}X>{\raggedright\arraybackslash}X>{\raggedright\arraybackslash}X>{\raggedright\arraybackslash}X>{\raggedright\arraybackslash}X@{}}
\hline
{\bfseries Controller} & {\bfseries Type} & {\bfseries Success} & {\bfseries 95\% CI} & {\bfseries Flip} & {\bfseries Gap viol.} & {\bfseries RMSE (m)} & {\bfseries Std (m)} & {\bfseries Mean XT (m)} & {\bfseries Max XT (m)} & {\bfseries Head. align.} & {\bfseries Action mag.} \\
\hline
PPO & RL & 100\% & --- & 0\% & 0\% & 0.146 & 0.140 & 1.238 & 1.908 & 0.935 & 0.330 \\
TD3 & RL & 0\% & 0--11\% & 0\% & 0\% & 0.055 & 0.052 & 1.218 & 1.908 & 0.987 & 0.138 \\
DDPG & RL & 63.3\% & 45.5--78.1\% & 30\% & 0\% & 0.375 & 0.239 & 1.489 & 1.908 & 0.475 & 0.758 \\
TRPO & RL & 100\% & --- & 0\% & 0\% & 0.142 & 0.140 & 1.224 & 1.908 & 0.833 & 0.230 \\
PID & Classical & 100\% & --- & 0\% & 0\% & 0.215 & 0.192 & 1.298 & 1.914 & 0.864 & 0.409 \\
SMC & Classical & 100\% & --- & 0\% & 0\% & 0.293 & 0.173 & 1.437 & 1.915 & 0.684 & 0.414 \\
Fuzzy & Classical & 100\% & --- & 0\% & 0\% & 0.243 & 0.149 & 1.392 & 1.914 & 0.834 & 0.400 \\
FL & Classical & 100\% & --- & 0\% & 0\% & 0.231 & 0.196 & 1.321 & 1.912 & 0.759 & 0.372 \\
MPC & Classical & 80\% & 62.7--90.5\% & 0\% & 6.67\% & 0.177 & 0.166 & 1.261 & 1.910 & 0.766 & 0.448 \\
ANFIS & Classical & 100\% & --- & 0\% & 0\% & 0.211 & 0.189 & 1.294 & 1.914 & 0.868 & 0.405 \\
\hline
\end{tabularx}
\par\vspace{3pt}\parbox{\textwidth}{\scriptsize \emph{SAC is excluded from Table XIX: its Coriolis-enabled evaluation artifacts were lost after a re-run of the training pipeline and were not reproduced in time for this comparison (Section XII-C). XT denotes cross-track error; head. align. is the mean cosine-similarity heading-alignment term of Section V-C; action mag. is the mean of the squared normalized thruster command over the episode (Section X-B).}}
\end{table*}

As shown in Figure 6 to 8, TD3 reports a 0 percent success rate alongside the lowest standoff RMSE of any controller in the table, 0.055 m. This apparent contradiction comes from a specific, reproducible failure mode. The TD3 checkpoint shows a corner-parking signature: it comes to rest exactly at the 5.00 m along-track corner waypoint and never proceeds around the route's single turn. The RMSE, mean cross-track error, and other tracking statistics in Table XIX are computed only over the part of the route the vehicle actually covers before it parks. A policy that tracks the first, straight half of the corridor almost perfectly and never attempts the second half therefore scores near-perfect local tracking and zero task completion at the same time. The two numbers measure different parts of what ``performance'' means for this controller, so they do not conflict. TD3 is the clearest case in Table XIX for reporting success rate and tracking error together, since either metric alone would give a misleading picture.

\begin{figure*}[!t]
\renewcommand{\thefigure}{6}
\centering
\includegraphics[width=0.9\textwidth,height=0.8\textheight,keepaspectratio]{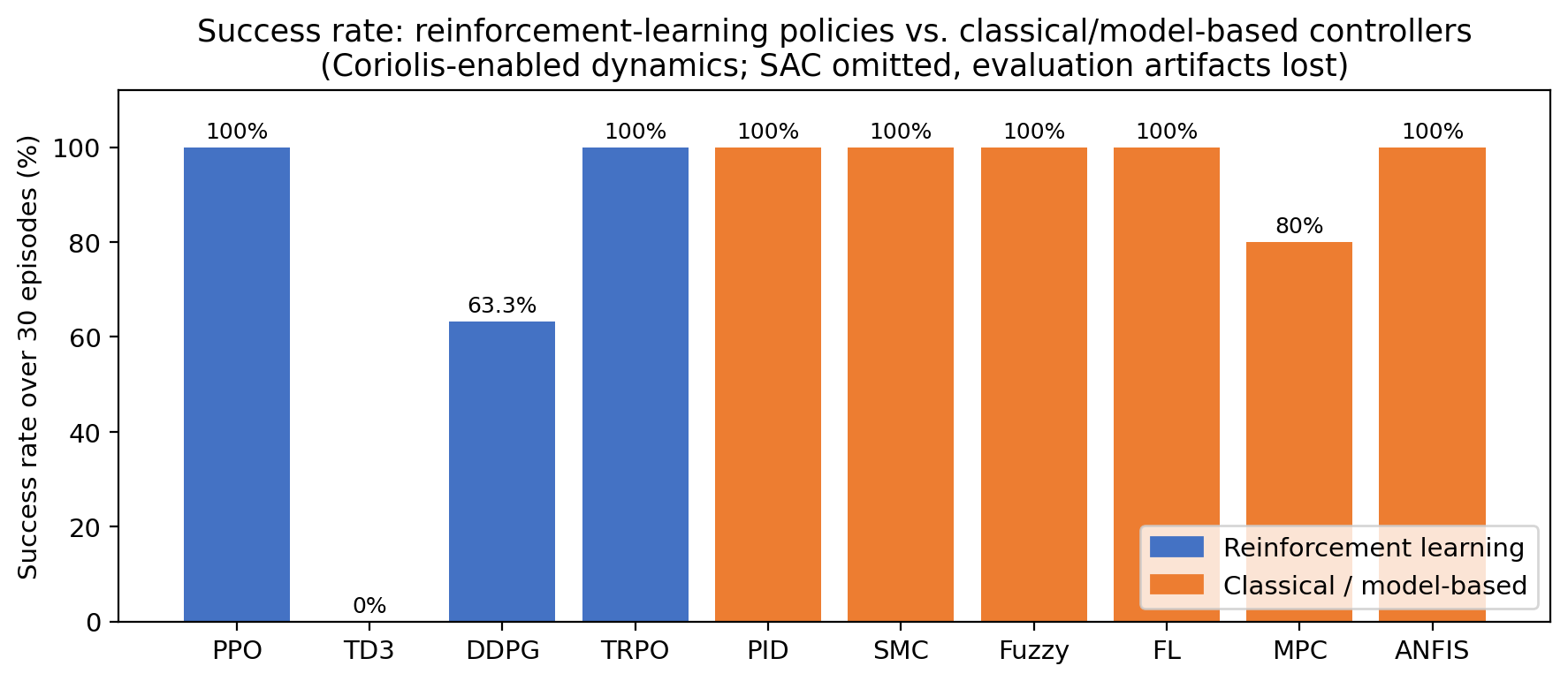}
\caption{Success rate for all ten evaluated controllers under Coriolis-enabled dynamics, reinforcement-learning policies and classical/model-based baselines colour-coded separately.}
\label{fig:6}
\end{figure*}

\begin{figure*}[!t]
\renewcommand{\thefigure}{7}
\centering
\includegraphics[width=0.9\textwidth,height=0.8\textheight,keepaspectratio]{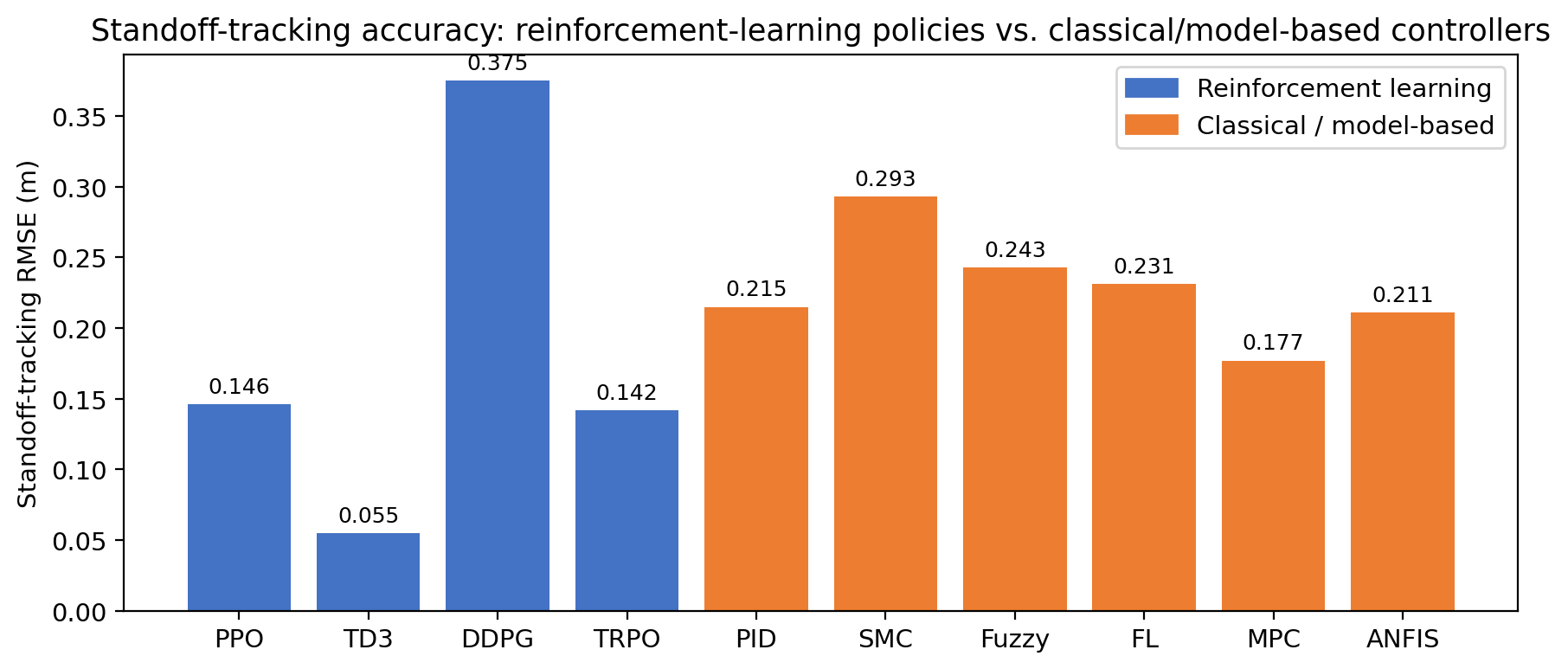}
\caption{Standoff root-mean-square tracking error for the same ten controllers.}
\label{fig:7}
\end{figure*}

\begin{figure*}[!t]
\renewcommand{\thefigure}{8}
\centering
\includegraphics[width=0.9\textwidth,height=0.8\textheight,keepaspectratio]{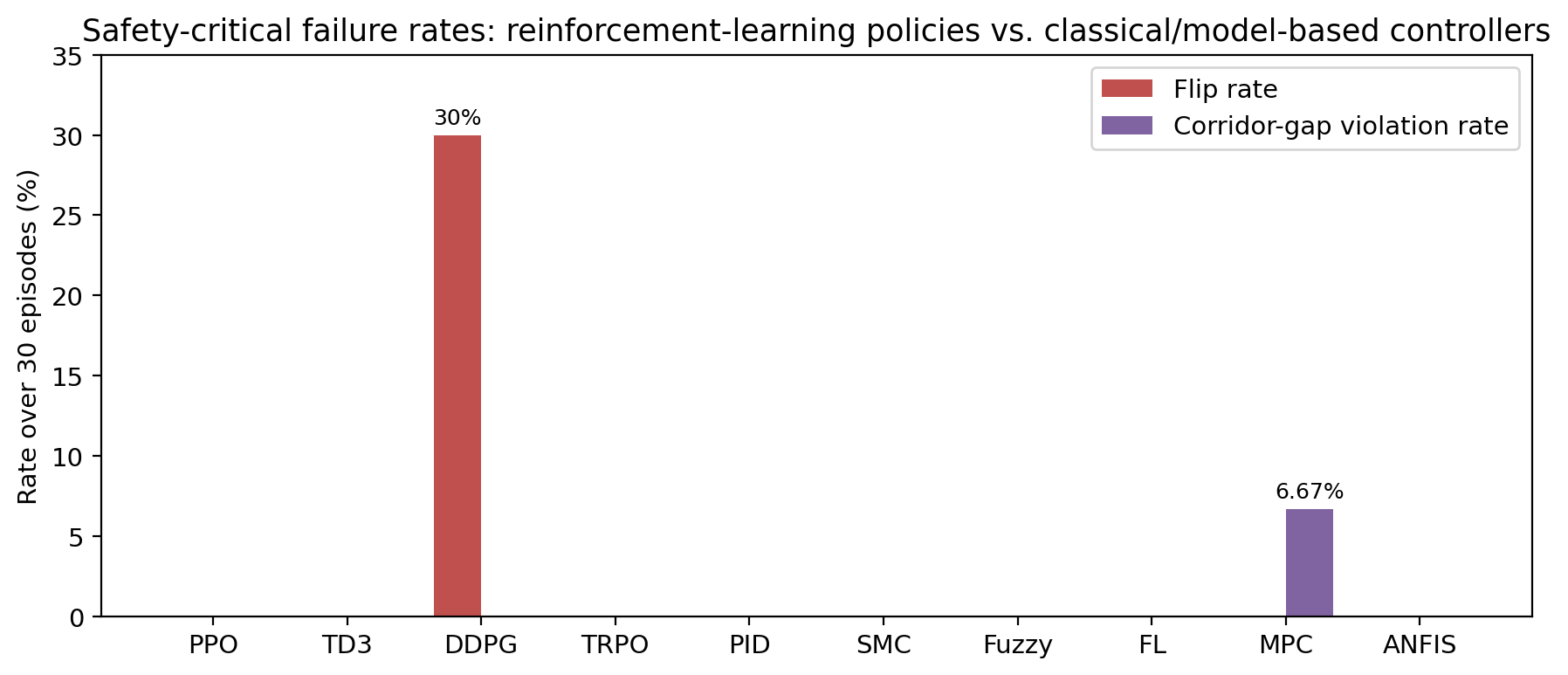}
\caption{Flip rate and corridor-gap-violation rate (safety metrics) for the same ten controllers.}
\label{fig:8}
\end{figure*}

Figures 9 through 14 give a further, trajectory-level view of the eight controllers that complete the route (Table XIX), drawn from each controller's canonical episode; TD3 and DDPG are omitted from these six figures because their episodes terminate, respectively, by corner-parking and by an early flip (Section XI-B), so neither produces a full-route trajectory comparable to the other eight.

\begin{figure*}[!t]
\renewcommand{\thefigure}{9}
\centering
\includegraphics[width=0.9\textwidth,height=0.8\textheight,keepaspectratio]{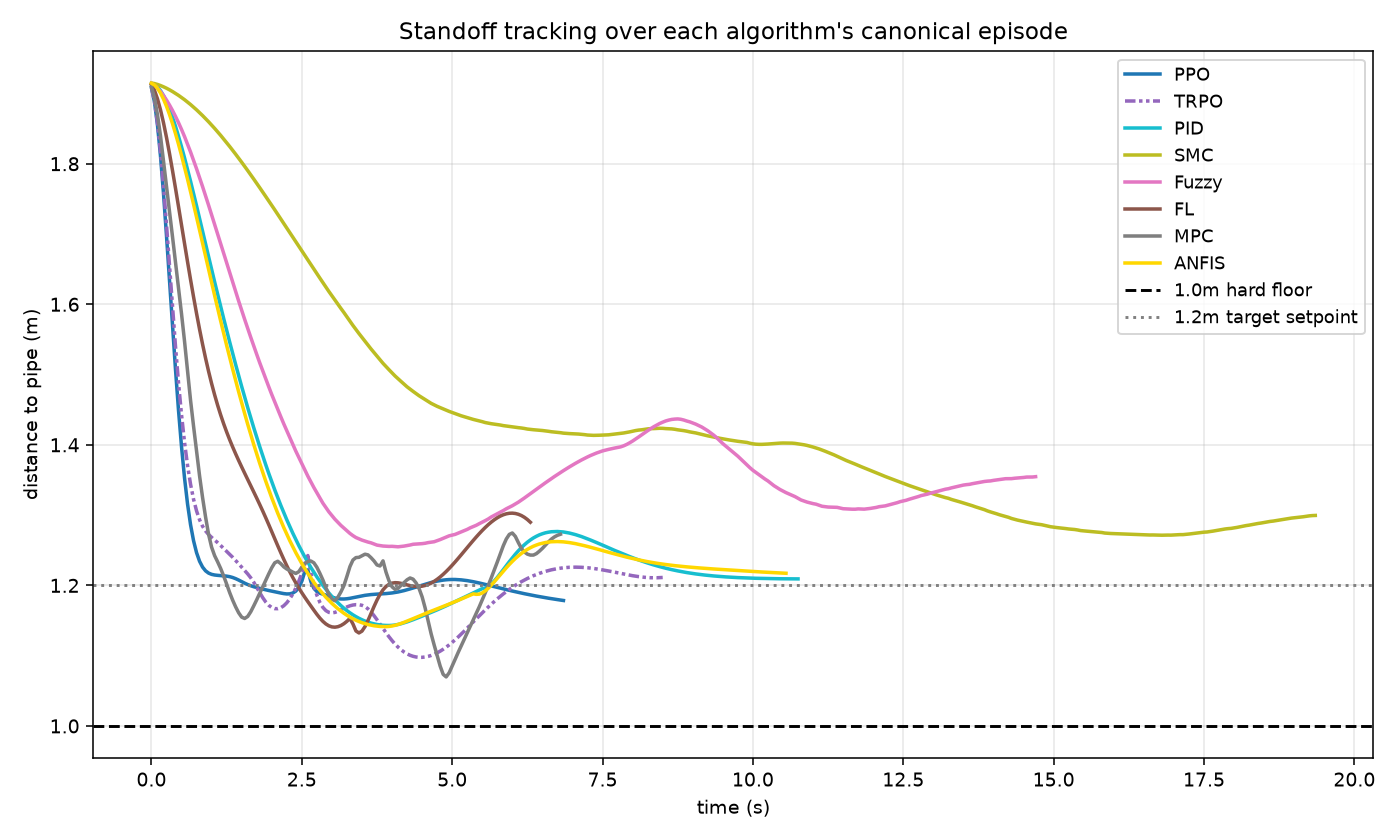}
\caption{Three-dimensional standoff distance to the pipeline versus time, canonical episode, for the eight controllers that complete the route, against the 1.2 m target setpoint and the 1.0 m hard floor of Table IX.}
\label{fig:9}
\end{figure*}

All eight curves in Fig. 9 start near the same initial standoff, just under 1.9 m, and converge toward the 1.2 m target without ever approaching the 1.0 m hard floor, consistent with the zero gap-violation rate every one of these eight controllers records in Table XIX. PPO converges fastest and settles closest to the setpoint, reaching the target band within about 2.5 s of the roughly 6.9 s episode; PID, TRPO, and ANFIS converge only slightly slower and settle with a small, quickly-damped oscillation. SMC and fuzzy logic control are the clear outliers: SMC's standoff distance plateaus around 1.4 m for several seconds before relaxing slowly toward the target over a much longer episode, and fuzzy logic control shows a visible secondary hump around 8-9 s before it, too, re-converges only in the final few seconds of its episode. This is a direct, visual confirmation of Table XIX's RMSE ranking: SMC and fuzzy logic control record the two highest standoff RMSE values among the eight successful controllers, 0.293 m and 0.243 m, and Fig. 9 shows why, a genuinely slower and less well-damped convergence rather than a brief transient. Feedback linearization and MPC each show a brief undershoot below the 1.2 m target, dipping to roughly 1.15 m around 4-5 s before correcting, consistent with a model-based controller occasionally over-correcting a transient error it can otherwise track tightly.

\begin{figure}[!t]
\renewcommand{\thefigure}{10}
\centering
\includegraphics[width=\columnwidth,height=0.8\textheight,keepaspectratio]{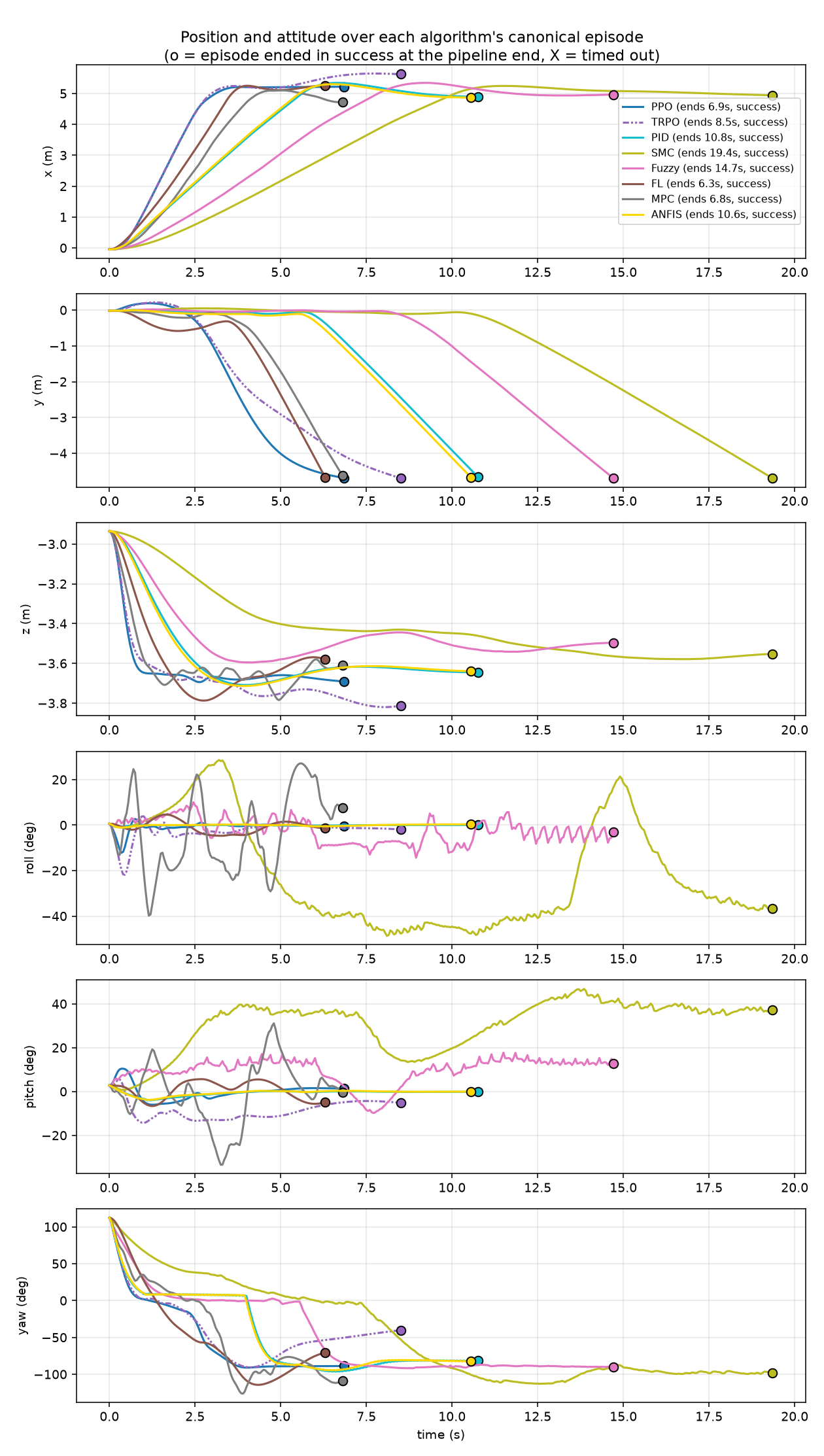}
\caption{Position (x, y, z) and attitude (roll, pitch, yaw) versus time, canonical episode, for the same eight controllers; a circular marker denotes success at the pipeline end.}
\label{fig:10}
\end{figure}

The x and y panels of Fig. 10 make each controller's episode length directly visible: PPO, feedback linearization, and MPC climb the along-track x-axis fastest and turn onto the final y-axis leg earliest, around t = 4 s, while SMC and fuzzy logic control climb far more gradually and do not begin the final leg until roughly t = 10-14 s, matching their markedly longer episode times in Table XX (19.4 s and 14.7 s, against 6.3-10.8 s for the other six). The z panel shows a shared initial dive from the -2.9 m spawn depth to the -3.6 to -3.8 m standoff band within the first two seconds for every controller, with TRPO diving deepest, to roughly -3.8 m, before settling back near -3.65 m. The roll and pitch panels distinguish the two model-free controllers most exposed to attitude disturbance: MPC shows the largest early attitude transient of any controller, roll and pitch each swinging roughly \ensuremath{\pm}30 degrees in the first five seconds, while SMC develops a sustained roll offset of roughly -40 degrees for much of the mid-episode before recovering near the end; fuzzy logic control shows smaller but persistent roll and pitch oscillation, of the order of \ensuremath{\pm}10-15 degrees, for almost the whole episode. The yaw panel shows all eight controllers turning from an initial heading near +110 degrees to roughly -80 to -100 degrees through the route's single corner, with SMC's yaw trace visibly the slowest and noisiest to settle, again consistent with it being the longest-running of the eight episodes.

\begin{figure}[!t]
\renewcommand{\thefigure}{11}
\centering
\includegraphics[width=\columnwidth,height=0.8\textheight,keepaspectratio]{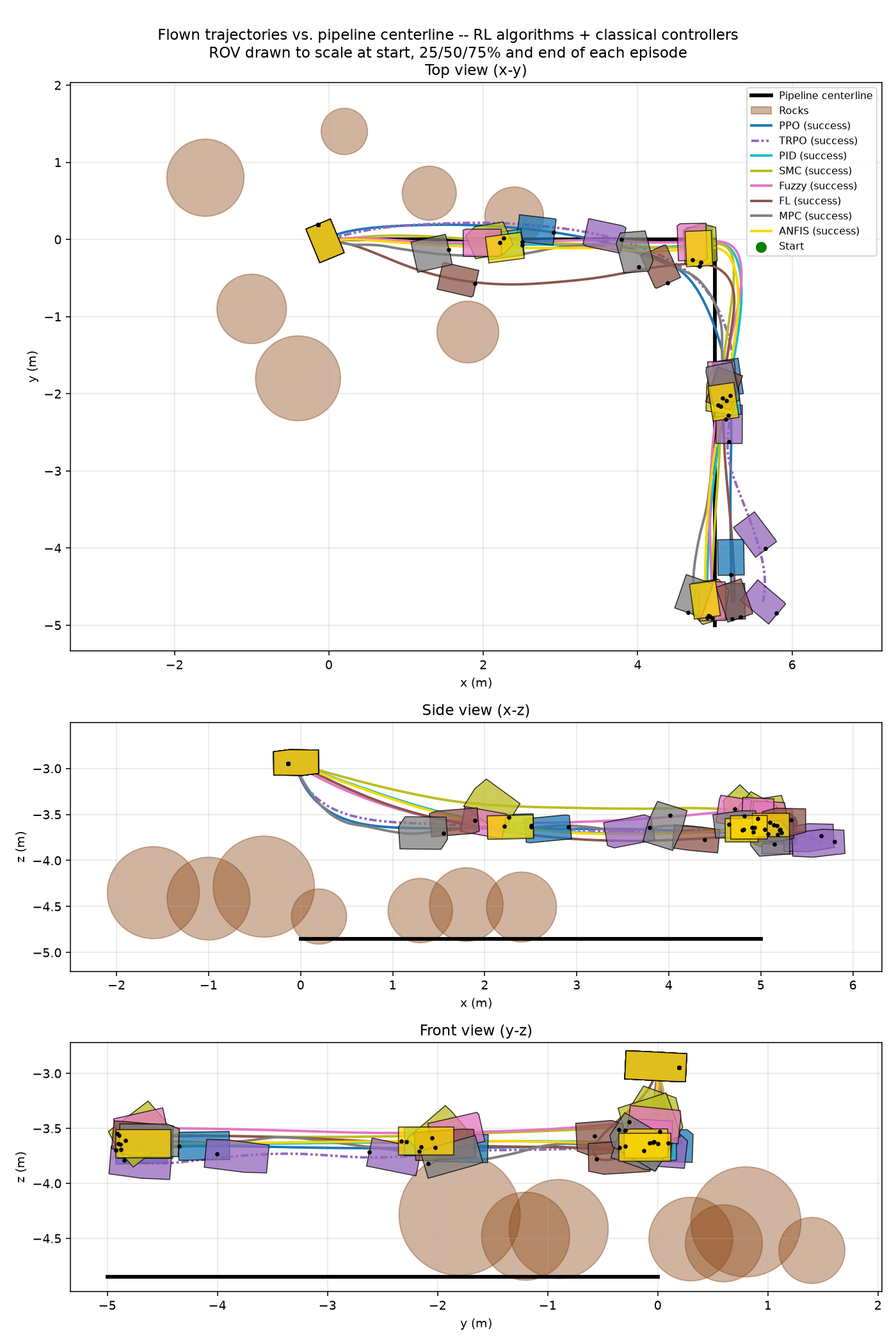}
\caption{Flown trajectories against the pipeline centerline, canonical episode, with the vehicle rendered to scale at the start, 25\%, 50\%, 75\%, and end waypoints: (a) top view (x-y), (b) side view (x-z), (c) front view (y-z).}
\label{fig:11}
\end{figure}

\begin{figure}[!t]
\renewcommand{\thefigure}{12}
\centering
\includegraphics[width=\columnwidth,height=0.8\textheight,keepaspectratio]{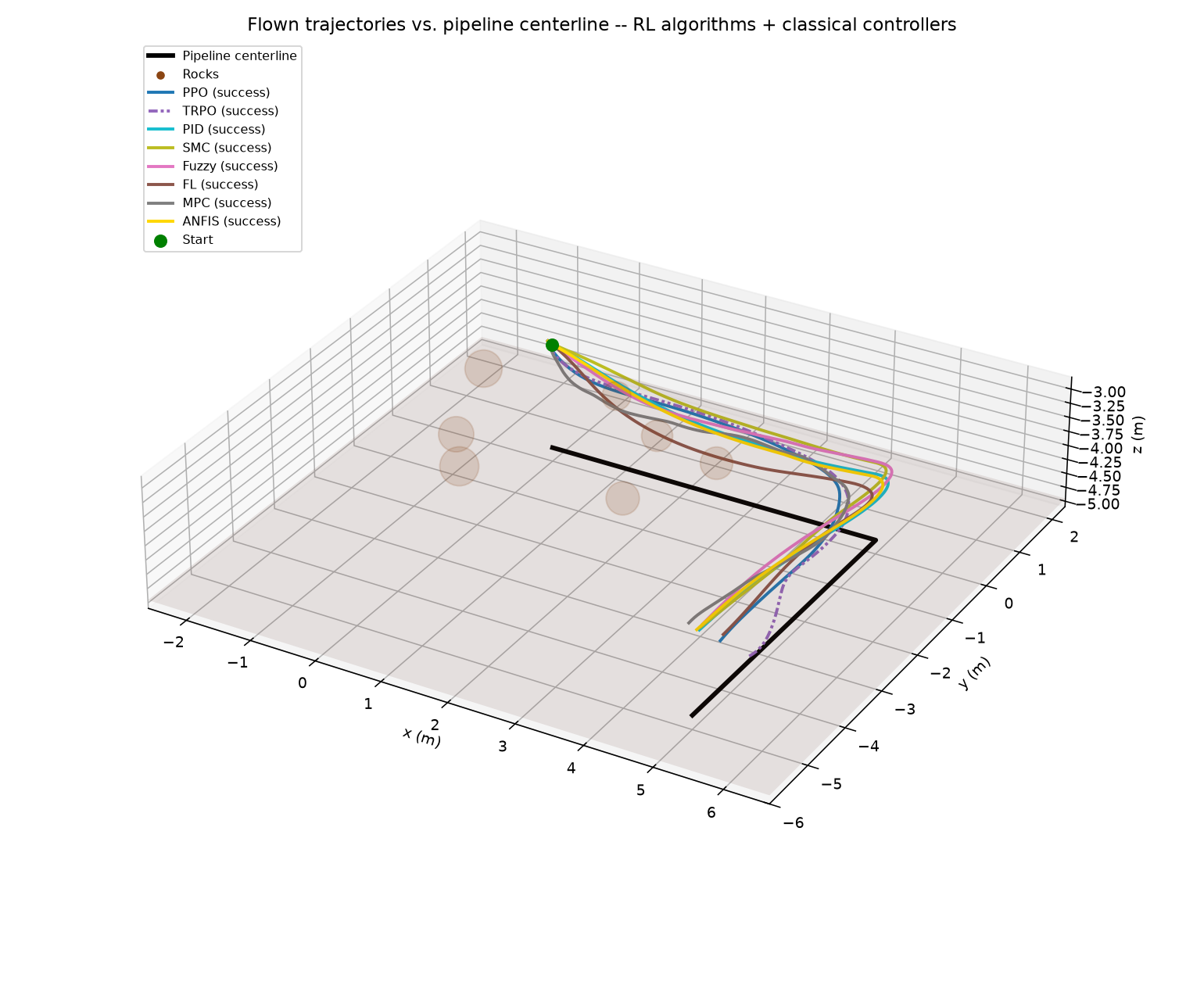}
\caption{The same flown trajectories of Fig. 11 in an oblique three-dimensional perspective.}
\label{fig:12}
\end{figure}

Figures 11 and 12 show all eight successful controllers holding a tight, near-identical corridor on both straight legs of the route, confirming that the single corner, not the straight segments, is where the eight controllers' behaviour actually separates. Feedback linearization is the clearest outlier in both figures, visibly swinging onto a wider arc before the corner and cutting back late rather than following the tighter line the other seven take; PID, SMC, fuzzy logic control, MPC, and ANFIS track within roughly one vehicle body-length of PPO and TRPO through the turn, consistent with the relatively narrow spread of standoff RMSE values, 0.146-0.293 m, these controllers record together in Table XIX. The side and front views in Fig. 11 show depth held within a narrow band throughout for every controller, with the small vertical spread visible at the corner, where lateral and vertical corrections briefly compound, the only visible departure from level flight.

\begin{figure}[!t]
\renewcommand{\thefigure}{13}
\centering
\includegraphics[width=\columnwidth,height=0.8\textheight,keepaspectratio]{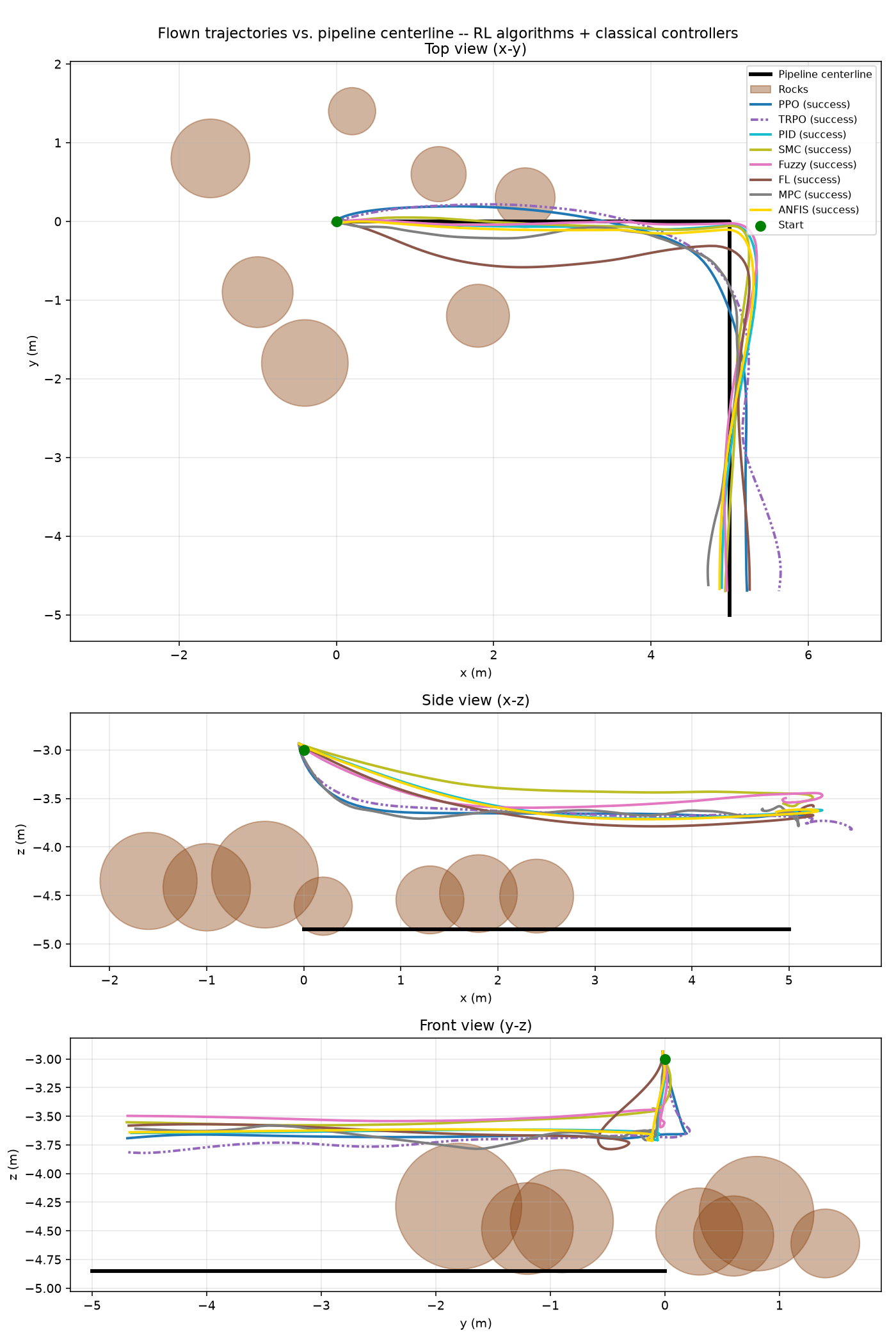}
\caption{The same eight flown trajectories as Fig. 11, plotted as centerlines only, over the full route from start to end.}
\label{fig:13}
\end{figure}

Fig. 13 drops the to-scale vehicle rendering of Fig. 11 and uses a wider view of the complete route instead. This wider framing reveals a small end-of-route correction in the side and front views for some controllers, as each settles within the 0.30 m waypoint radius of Section V-A. Fig. 11 is cropped more tightly and ends at the final waypoint marker, so this last adjustment does not appear there.

\begin{figure}[!t]
\renewcommand{\thefigure}{14}
\centering
\includegraphics[width=\columnwidth,height=0.8\textheight,keepaspectratio]{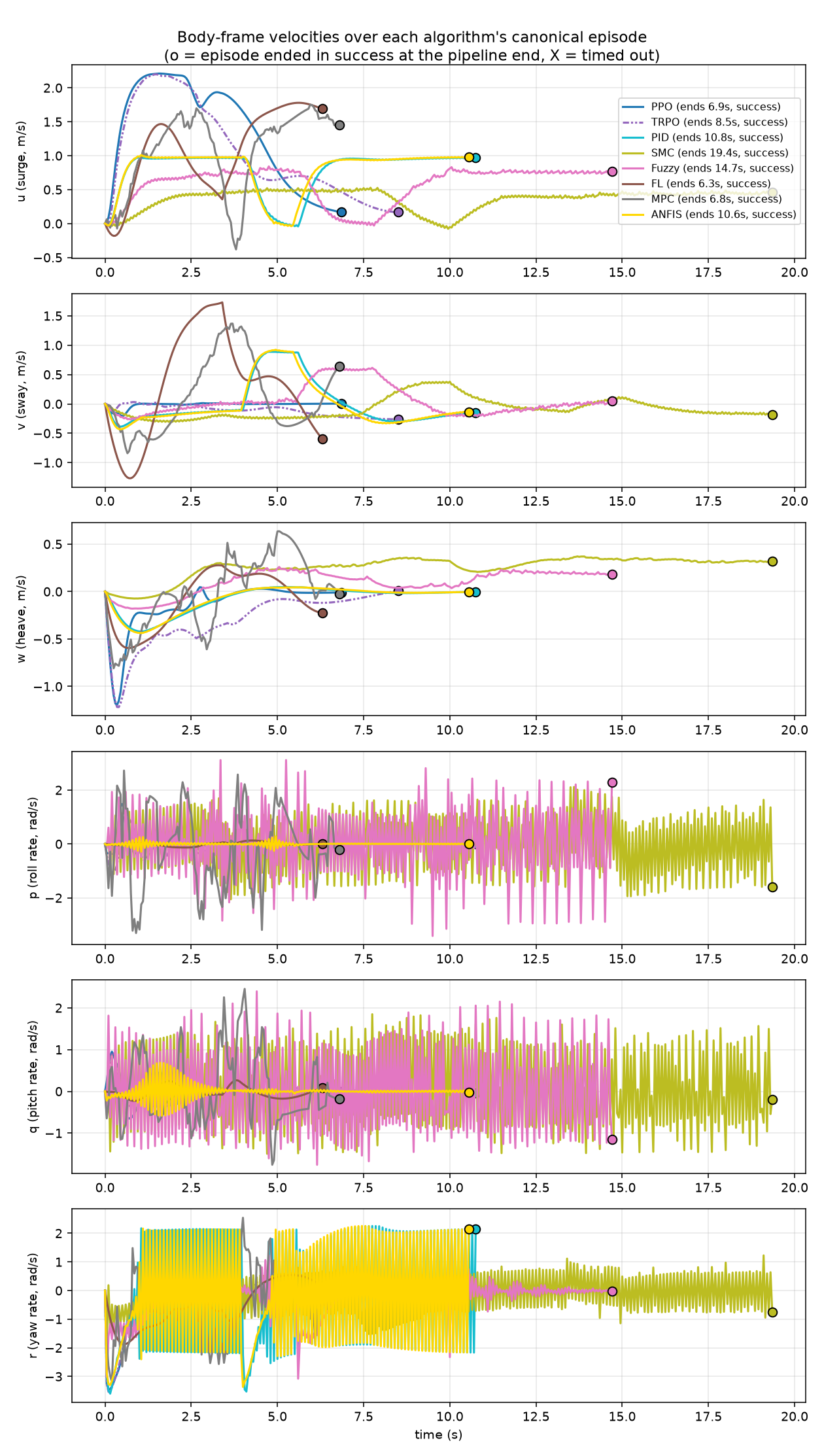}
\caption{Body-frame linear velocity (surge u, sway v, heave w) and angular velocity (roll rate p, pitch rate q, yaw rate r) versus time, canonical episode, for the same eight controllers.}
\label{fig:14}
\end{figure}

The surge panel of Fig. 14 shows PPO and feedback linearization reaching the highest peak surge velocity, 2.0-2.2 m/s, during the initial acceleration before settling toward the 1.825 m/s cruise setpoint of Table IX, while SMC and MPC show a visibly lower and noisier surge profile that oscillates for most of the episode; PID and ANFIS converge cleanly to a steady surge near 1.0 m/s by around t = 5 s. The sway and heave panels show a brief, sharp transient in the first one to two seconds, as each controller turns into the corridor and settles to its cruise depth, with feedback linearization and MPC showing the largest initial sway and heave-velocity excursions of the eight. The roll-rate and pitch-rate panels give the clearest evidence in this figure set of sustained, high-frequency command activity: fuzzy logic control and SMC both show a dense, continuously oscillating band across nearly the whole episode, in visible contrast with PPO, TRPO, PID, feedback linearization, MPC, and ANFIS, which settle near zero roll and pitch rate after a brief initial transient. This is a direct illustration of the chatter the boundary-layer reaching law of Section VIII-B is specifically designed to reduce in SMC's thruster commands rather than its attitude-rate response, and shows that the hand-designed fuzzy controller of Section VIII-C exhibits a broadly similar high-frequency roll/pitch-rate signature despite using a smooth, weighted-average output rule rather than a switching law. The yaw-rate panel shows PID and ANFIS sharing a closely similar, sustained oscillation band from roughly t = 5 s to the end of the episode, a visible signature consistent with ANFIS being trained to behaviour-clone PID's own error-to-output mapping (Section VIII-F) rather than independently optimized against the tracking task.

Table XX reports additional training-cost, energy, and convergence indicators for the four reinforcement learning policies carried into Table XIX's main comparison, computed from the canonical episode and the training logs of Sections VI and VII rather than from the 30-episode randomized evaluation.

\begin{table*}[!t]
\renewcommand{\thetable}{XX}
\caption{Additional quantitative indicators, canonical episode and training logs (PPO, TD3, DDPG, TRPO)}
\label{tab:XX}
\centering\footnotesize
\setlength{\tabcolsep}{3pt}
\begin{tabularx}{\textwidth}{@{}>{\raggedright\arraybackslash}X>{\raggedright\arraybackslash}X>{\raggedright\arraybackslash}X>{\raggedright\arraybackslash}X>{\raggedright\arraybackslash}X@{}}
\hline
{\bfseries Metric} & {\bfseries PPO} & {\bfseries TD3} & {\bfseries DDPG} & {\bfseries TRPO} \\
\hline
Episode time (s) & 6.9 & 52.0 & 1.1 & 8.6 \\
Along-track progress (of 10 m route) & 9.70 m & 5.00 m & 0.38 m & 9.71 m \\
Mean \textbar thrust\textbar{} (normalized) & 0.330 & 0.138 & 0.758 & 0.230 \\
Energy (normalized thrust\ensuremath{^{2}} \ensuremath{\cdot} s) & 11.8 & 15.7 & 5.9 & 9.4 \\
Energy per metre travelled & 1.2 & 3.1 & n/a (negligible progress) & 1.0 \\
Steps to 90\% deterministic success & 10,485,760 & never reached & never reached & 14,745,600 \\
Wall-clock to 90\% deterministic success (min) & 4.9 & never reached & never reached & 6.9 \\
Total training timesteps (actual) & 20,004,864 & 8,994,816 & 2,998,272 & 20,004,864 \\
Training timestep budget & 20,000,000 & 9,000,000 & 3,000,000 & 20,000,000 \\
Training wall-clock time (min) & 9.2 & 74.5 & 28.6 & 9.4 \\
Environment throughput (steps/s) & 36,121 & 2,013 & 1,747 & 35,428 \\
Actor network parameters & 74,504 & 74,504 & 74,504 & 74,504 \\
\hline
\end{tabularx}
\par\vspace{3pt}\parbox{\textwidth}{\scriptsize *Energy is the sum, over the canonical episode, of the squared normalized thruster command across all eight thrusters, a shape-independent proxy for actuation cost rather than a measurement in joules (Section X-B); a timed-out episode, such as TD3's, accumulates this proxy over its full duration, so energy per metre is reported only for controllers that make progress along the route. Steps and wall-clock time to 90\% deterministic success are read from the first evaluation checkpoint during training whose deterministic policy reaches at least 90\% success (Section X-B); DDPG never crosses this threshold despite its 63.3\% success rate in the 30-episode randomized evaluation of Table XIX, since that figure is measured under domain-randomized rock layouts distinct from the fixed-layout deterministic checkpoints logged during training. Actor network parameter counts are identical across all four policies by construction (Section VII), confirming that the differences reported elsewhere in this table and in Table XIX reflect the algorithm and its training dynamics rather than any difference in model capacity.}
\end{table*}

Table XX makes explicit a point implicit in Table XIX's success and flip columns: TD3's very low mean thrust magnitude, 0.138, the smallest of the four, still yields the single highest energy figure, 15.7, because its 52 s timed-out episode is more than five times longer than PPO's; a controller that commands gentle thruster forces for a long time can spend more actuation effort overall than one that commands larger forces briefly, so mean action magnitude and total energy are informative in combination rather than interchangeably. DDPG's reported energy, 5.9, is the lowest of the four despite its far larger mean thrust magnitude, 0.758, because its canonical episode terminates by flip after only 1.1 s (Table XIX), too briefly for its high per-step actuation cost to accumulate; the ``n/a'' entry under energy per metre reflects that this episode made negligible along-track progress before terminating; this is a further illustration, alongside TD3's corner-parking (Section XI-B), of why a single summary number can mislead when read in isolation from the episode outcome it was computed over. PPO and TRPO's roughly 20-fold throughput advantage over TD3 and DDPG (35,000-36,000 against 1,700-2,000 environment-steps per second) is consistent with the on-policy/off-policy distinction discussed in Section VII-F: the on-policy rollouts parallelize cleanly across 64 environments with no gradient step between environment steps, while the off-policy trainers perform eight gradient updates per environment step (Section VII-A), which dominates their wall-clock cost despite their much smaller total-timestep budgets.

Table XXI consolidates, for the same four policies, the headline optimization hyperparameters already given individually in Tables VII and VIII (Sections VI-D and VII-F) alongside each algorithm's own distinguishing settings, so the four controllers compared side by side in Table XIX can also be compared side by side on how they were trained.

\begin{table*}[!t]
\renewcommand{\thetable}{XXI}
\caption{Consolidated hyperparameters and algorithm-specific settings (PPO, TD3, DDPG, TRPO)}
\label{tab:XXI}
\centering\footnotesize
\setlength{\tabcolsep}{3pt}
\begin{tabularx}{\textwidth}{@{}>{\raggedright\arraybackslash}X>{\raggedright\arraybackslash}X>{\raggedright\arraybackslash}X>{\raggedright\arraybackslash}X>{\raggedright\arraybackslash}X@{}}
\hline
{\bfseries Setting} & {\bfseries PPO} & {\bfseries TD3} & {\bfseries DDPG} & {\bfseries TRPO} \\
\hline
Learning rate & 3\ensuremath{\times}10\ensuremath{^{-4}} (actor and critic, Adam) & 3\ensuremath{\times}10\ensuremath{^{-4}} (actor and critic, Adam) & 3\ensuremath{\times}10\ensuremath{^{-4}} (actor and critic, Adam) & Policy: none (natural gradient); value 1\ensuremath{\times}10\ensuremath{^{-3}} (Adam) \\
Discount factor, \ensuremath{\gamma} & 0.99 (GAE \ensuremath{\lambda} = 0.95) & 0.99 & 0.99 & 0.95 (GAE \ensuremath{\lambda} = 0.95) \\
Clipping / trust-region parameter & Clip ratio 0.2 & Target-noise clip 0.5 (Polyak \ensuremath{\tau} = 0.005) & Polyak \ensuremath{\tau} = 0.005 & Max KL 0.01 (trust region), no ratio clip \\
Batch size & 4,096 minibatch (16,384-sample rollout: 64 environments \ensuremath{\times} 256 steps) & 256 (replay buffer 1M) & 256 (replay buffer 1M) & 16,384 full rollout (value minibatch 4,096) \\
Updates per rollout / step & 10 epochs per rollout & 8 gradient updates per environment step & 8 gradient updates per environment step & 1 conjugate-gradient step (10 CG iterations) + 10 value epochs \\
Network & MLP, 2\ensuremath{\times}256, tanh (actor and critic) & MLP, 2\ensuremath{\times}256, tanh (actor, twin critics) & MLP, 2\ensuremath{\times}256, tanh (actor, critic) & MLP, 2\ensuremath{\times}256, tanh (actor and critic) \\
Optimizer & Adam & Adam & Adam & Conjugate gradient + line search (policy); Adam (value) \\
Algorithm-specific setting & Entropy coefficient 0.01; value-function coefficient 0.5; max gradient norm 0.5 & Exploration noise \ensuremath{\sigma} = 0.20; target-policy smoothing noise \ensuremath{\sigma} = 0.20; policy-update delay 2 & Exploration noise \ensuremath{\sigma} = 0.10 (\ensuremath{\sigma} = 0.20 was tried and degraded performance) & Fisher damping 0.1; backtracking line search; action-penalty coefficient 0.01 (live-flight roll/pitch damping) \\
\hline
\end{tabularx}
\par\vspace{3pt}\parbox{\textwidth}{\scriptsize \emph{This table restates, side by side, hyperparameters individually reported for PPO in Table VII (Section VI-D) and for TD3, DDPG, and TRPO in Table VIII (Section VII-F); it adds no new numeric values, but aligns them by row so the four controllers appearing together in Table XIX can be compared on training configuration as directly as they are compared on outcome.}}
\end{table*}

\subsection{Other findings}

The corner-cutting projection exploit found during environment development (Section V-E) comes from the guidance geometry and affects every controller alike. An early, unguarded projection of the vehicle's position onto the piecewise-linear pipeline centerline allowed a diagonal path to collect progress reward by cutting across the outside of the route's corner without ever approaching the true corner vertex. Two fixes close this gap: the per-segment projection gate of Section V-E and the look-ahead heading point described in the closing paragraph of Section VIII. Every controller compared in Table XIX benefits from both equally.

DDPG's 30 percent flip rate in Table XIX is consistent with its action-magnitude value of 0.758, roughly two to three times the 0.230 to 0.448 range of every other controller in the table. A policy that commands much larger thruster forces on average is also the one most prone to the large, sudden attitude changes that end in a flip. The Discussion returns to this connection in Section XII-B.

\section{Discussion}

\subsection{What the Coriolis-term ablation shows}

The ablation of Section XI-A shows that PPO and TRPO both handle the added nonlinear coupling introduced by the Coriolis-centripetal term well. Neither algorithm loses any success or safety performance when the term is included, and both track more accurately with it, with the clearest gain for TRPO. For the two on-policy algorithms and the reward and observation design used in this study, these results support modelling the Coriolis term by default. Only PPO and TRPO were formally compared in this ablation (Section XI-A), so extending the finding to off-policy algorithms such as SAC, TD3, and DDPG is a natural next step for the multi-algorithm follow-up noted in Section XII-D.

\subsection{Reinforcement learning versus classical and model-based control}

Table XIX's main comparison shows that model-based feedback linearization and the two on-policy/trust-region reinforcement-learning policies, PPO and TRPO, achieve the strongest combination of a 100 percent success rate and competitive-to-best tracking accuracy, 0.146 m, 0.142 m, and 0.231 m RMSE respectively. PID, fuzzy logic control, and ANFIS all also reach a 100 percent success rate but with worse tracking accuracy, in roughly that order, PID best of the three at 0.215 m, ANFIS close behind at 0.211 m, and fuzzy logic control at 0.243 m; sliding-mode control reaches 100 percent success but the worst tracking accuracy of any successful controller, 0.293 m, consistent with the boundary-layer reaching law of Section VIII-B trading tracking precision for smooth, chatter-free thruster commands. Model predictive control trades some success rate, 80 percent, for RMSE, 0.177 m, competitive with PID and close to PPO and TRPO; a shooting-based planner limited to a 20-step horizon and two cross-entropy-method iterations (Section VIII-E) can produce an occasional short-sighted or locally infeasible plan around the route's single corner without enough horizon or refinement to recover from it within an episode, plausibly the source of MPC's non-zero corridor-gap-violation rate, the only classical controller with one.

Among the reinforcement-learning policies, PPO and TRPO clearly dominate DDPG and TD3 under this reward and training budget. DDPG's instability, a 30 percent flip rate and an action magnitude roughly two to three times every other controller's (Section XI-C), is consistent with the known brittleness of a single-critic deterministic policy gradient under Q-overestimation; TD3's corner-parking (Section XI-B) is a specific, reproducible degenerate local optimum tied to this checkpoint's exploration-noise and discount-factor setup, not evidence that off-policy methods generally fail this task, since TD3 tracks the portion of the route it does traverse more tightly than any other controller in the comparison.

The broader synthesis is that classical and model-based control, with no learning involved at all for PID and feedback linearization, and comparatively little for the hand-designed fuzzy controller, remains highly competitive with the best reinforcement-learning policies on this task. Feedback linearization in particular, which has privileged access to the true dynamics model, matches or exceeds the tracking accuracy of PPO and TRPO without any training cost at all, an honest and important caveat to any narrative in which reinforcement learning is assumed to outperform classical control outright. Reinforcement learning's advantage here is less about outright dominance on this metric set and more about not requiring a hand-derived dynamics model, a gain-tuning process, or a rule base, at the cost of sensitivity to algorithm choice and training budget, the specific TD3 and DDPG failure modes above being the clearest illustration of that cost.

Lastly, the MPC baseline uses a deliberately lightweight planning budget (Section VIII-E), so its 80 percent success rate may understate what shooting-based MPC can achieve. The ANFIS baseline (Section VIII-F) is a learned clone of PID and was not optimized independently. All experiments use a single L-shaped route of roughly 10 m with narrow rock randomization, and each controller comes from a single run. Both the Coriolis-term ablation and the classical-control comparison show the value of testing a modelling or algorithm choice through controlled comparison on the same evaluation harness. Neither the safety of a simplification nor the advantage of a learned policy over a hand-derived one should be assumed in advance. Several extensions follow directly from this study. The Coriolis ablation of Section XI-A can be extended to the three off-policy algorithms not formally compared here. MPC can be tested with a longer horizon and a more refined planning budget, and ANFIS can be tuned independently of PID. System-identified hydrodynamic parameters from a physical vehicle and a multi-seed statistical treatment across all controllers would further strengthen the results. Finally, the same environment, reward, and observation definitions should transfer directly to a training loop built on Isaac Lab (Section II-E), replacing the purpose-written NumPy trainer used here and allowing the parallel-environment count to scale well beyond the current 64 instances.

\section{Conclusion}

This paper presented a pipeline-tracking control architecture for a six-degree-of-freedom ROV, in which a single USD scene supplies Coriolis-coupled Fossen dynamics parameters to both a vectorized training model and an interactive NVIDIA Isaac Sim deployment. A controlled ablation over PPO and TRPO validated the inclusion of the Coriolis-centripetal term. The shared-dynamics pipeline was then extended with six classical and model-based baselines (PID, sliding-mode control, fuzzy logic control, feedback linearization, model predictive control, and an adaptive neuro-fuzzy inference system), evaluated under identical guidance geometry and thruster allocation alongside four reinforcement learning algorithms: PPO, TD3, DDPG, and TRPO. Including the Coriolis term leaves PPO and TRPO stable and improves their tracking accuracy, most clearly for TRPO, which supports modelling it by default for this task family. In the live Isaac Sim controller, only the added-mass component is applied explicitly, since the PhysX solver already supplies the rigid-body component. Under Coriolis-enabled dynamics and a shared 30-episode evaluation protocol with Wilson confidence intervals, PPO, TRPO, and feedback linearization perform best overall, combining a 100 percent success rate with the highest tracking accuracy in the comparison. DDPG and TD3 each show a specific, explainable failure mode: DDPG learns an unstable, high-action-magnitude policy, and TD3 settles into a reproducible corner-parking local optimum. Both are traceable to the individual policies and do not indicate a general weakness of off-policy methods. Classical and model-based control, PID and feedback linearization in particular, remains a strong and comparatively low-cost baseline for this task family, and the results show that hand-derived control can match learned policies on this problem. The findings are bounded by a deliberately lightweight MPC planning budget, an ANFIS baseline that clones PID, single-seed training and tuning runs, the absence of an ocean-current term, and the lack of physical-vehicle validation.

\section*{Acknowledgment}

The author acknowledges the NVIDIA-Newcastle Joint Laboratory (NVAITC) at Newcastle University Singapore for access to the hardware and software used to run NVIDIA Isaac Sim and the reinforcement learning experiments in this study: an NVIDIA GB10 system with a 20-core ARM (aarch64) CPU (10 Cortex-X925 cores up to 3.9 GHz and 10 Cortex-A725 cores up to 2.8 GHz), an NVIDIA GB10 GPU with unified memory shared with system RAM (driver 580.173.02), 121 GiB of system memory, and a 3.7 TB NVMe solid-state root filesystem, running Ubuntu 24.04 LTS (Linux kernel 6.17) with Python 3.12, NVIDIA Isaac Sim 6.0.1.0, PyTorch 2.11.0 (CUDA 13), NumPy 2.3.1, SciPy 1.17.0, Matplotlib 3.10.8, and Pillow 12.2.0.

\section*{Data Availability Statement}

The trained policy checkpoints for all five compared reinforcement-learning algorithms, the tuned parameters of the six classical and model-based controllers, the USD scene, and the evaluation logs generated during this study are available from the corresponding author on reasonable request.

\section*{Declaration of Competing Interest}

The author declares no known competing financial interests or personal relationships that could have appeared to influence the work reported in this paper.

\section*{Ethics Statement}

This study involved simulation of an underwater vehicle and did not involve human or animal subjects; no ethics approval was required.

\begin{IEEEbiography}[{\includegraphics[width=1in,height=1.25in,clip,keepaspectratio]{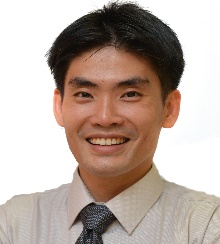}}]{Cheng Siong Chin}. He received his Ph.D in Applied Control Engineering at Research Robotics Centre in Nanyang Technological University (NTU) in 2008 and M.Sc. (Distinction) in Advanced Control and Systems Engineering from The University of Manchester in 2001. He also graduated with a B.Eng (Hons) in Mechanical and Aerospace Engineering from NTU in 2000.

He is currently a Chair Professor of Intelligent Systems Modeling and Simulation at Newcastle University, where he also serves as the Director of Research and Innovation. He was a Lecturer in Mechatronics Engineering at Temasek Polytechnic. Before moving into academia, he worked in the consumer electronics industry for a few years. He currently holds 3 U.S. Patents in electronics test systems and components. His research interests encompass the design and simulation of intelligent systems in uncertain environments, with applications in areas such as marine electric vehicles, energy storage systems, and acoustic systems. He serves as the Editor-in-Chief (EiC) for Cybernetics and Systems and as an Associate Editor for both IEEE Access and Applied Artificial Intelligence. Additionally, he has been a General Chair and a Keynote Speaker at various international conferences.
\end{IEEEbiography}

\begin{IEEEbiography}[{\includegraphics[width=1in,height=1.25in,clip,keepaspectratio]{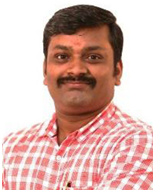}}]{M. Venkateshkumar}. Mohan Venkateshkumar (Member, IEEE) is currently an Assistant Professor (SL Grade) with the Department of Electrical and Electronics Engineering, Amrita School of Engineering, Amrita Vishwa Vidyapeetham, Coimbatore Campus. With more than 15 years of teaching and research experience, he specializes in power systems engineering. He has conducted significant research in the field of design and control of artificial intelligence-based hybrid renewable energy sources for both on-grid and off-grid applications, as well as electric vehicles. He has published more than 60 research papers in international journals and conferences and serves as a technical committee member for more than 40 international journals and conferences. He has authored three books/chapters with Elsevier and Intech open publications. In addition, he is a member of IEEE P825 and P2030 research working groups in USA. He received the Young Scientist Award from the government of Tamil Nadu. In addition, he has been recognized for his outstanding contributions to the field by receiving the MATLAB Outstanding Contribution Award for three consecutive years, from 2015 to 2017. He holds the position of the Vice Chair of the IEEE Power and Energy Society and the IEEE Education Society IEEE Madras Section. He is an Associate Editor of IEEE Access journal, which has an impact factor of 4.09
\end{IEEEbiography}

\begin{IEEEbiography}[{\includegraphics[width=1in,height=1.25in,clip,keepaspectratio]{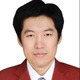}}]{Jianhua Zhang}. Prof. Dr. Jianhua Zhang is an Associate Professor and master's supervisor at the School of Information and Control Engineering, Qingdao University of Technology. He graduated from Yanshan University in 2011 with a Ph.D. in Control Theory and Control Engineering and a postdoctoral fellowship from the University of the West of England, UK. He is the editor-in-chief of the International Journal of Applied Mathematics in Control Engineering and the associate editor of the International Journal of Analytical Methodologies and Applications in Engineering. He is an associate editor of the International Journal of Analytical Methodologies and Applications in Engineering, and a member of the international program committee of the International Conference of Modelling, Identification and Control. His main research interests are neural network adaptive control and sliding mode control. He has published more than 50 papers, including more than 30 EI/SCI papers, 3 academic monographs, and 2 patents for inventions and utility models.
\end{IEEEbiography}

\end{document}